\documentclass[sigconf,natbib=True]{acmart}
\AtBeginDocument{%
  \providecommand\BibTeX{{%
    \normalfont B\kern-0.5em{\scshape i\kern-0.25em b}\kern-0.8em\TeX}}}

\copyrightyear{2024}
\acmYear{2024}
\setcopyright{acmlicensed}
\acmConference[RecSys '24]{18th ACM Conference
on Recommender Systems}{October 14--18, 2024}{Bari, Italy}
\acmBooktitle{18th ACM Conference on Recommender Systems (RecSys '24),
October 14--18, 2024, Bari, Italy}
\acmDOI{10.1145/3640457.3688142}
\acmISBN{979-8-4007-0505-2/24/1 }

\usepackage{multirow}
\usepackage{amsthm}
\usepackage{tocloft} 
\usepackage{bbding}

\usepackage{subfig}

\makeatletter
\ifdefined\checkmark
  \let\checkmark\@undefined
\fi
\makeatother

\usepackage{dingbat} 
\newcommand{\algname}{PEBOL}
\newlistof{example}{loe}{List of Examples}

\begin{document}

\title{Bayesian Optimization with LLM-Based Acquisition Functions for Natural Language Preference Elicitation}

\newcommand{\anton}[1]{{\color{red}[Anton: #1]}}
\newcommand{\scott}[1]{{\color{green}[Scott: #1]}}
\newcommand{\david}[1]{{\color{purple}[David: #1]}}
\setlength{\abovecaptionskip}{5pt}
\setlength{\belowcaptionskip}{0pt}

\author{David Eric Austin}
\authornote{Both authors contributed equally to this research.}
\email{deaustin@uwaterloo.ca}
\affiliation{%
  \institution{University of Waterloo}
  \city{Waterloo}
  \state{Ontario}
  \country{Canada}
}

\author{Anton Korikov}
\authornotemark[1]
\email{anton.korikov@mail.utoronto.ca}
\author{Armin Toroghi}
\author{Scott Sanner}
\affiliation{%
  \institution{University of Toronto}
  \city{Toronto}
  \state{Ontario}
  \country{Canada}
}

\renewcommand{\shortauthors}{Austin and Korikov, et al.}

\begin{abstract}
Designing preference elicitation (PE) methodologies that can quickly ascertain a user's top item preferences in a cold-start setting is a key challenge for building effective and personalized conversational recommendation (ConvRec) systems. While large language models (LLMs) enable fully natural language (NL) PE dialogues, we hypothesize that monolithic LLM NL-PE approaches lack the multi-turn, decision-theoretic reasoning required to effectively balance the exploration and exploitation of user preferences towards an arbitrary item set. In contrast, traditional Bayesian optimization PE methods define theoretically optimal PE strategies, but cannot generate arbitrary NL queries or reason over content in NL item descriptions -- 
requiring users to express preferences via ratings or comparisons of unfamiliar items. To overcome the limitations of both approaches, 
we formulate NL-PE in a Bayesian Optimization (BO) framework that seeks to actively elicit NL feedback 
to identify the best recommendation. Key challenges in generalizing BO to deal with natural language feedback include determining: (a) how to leverage LLMs to model the likelihood of NL preference feedback as a function of item utilities, and (b) how to design an acquisition function for NL BO that can elicit preferences in the infinite space of language.
We demonstrate our framework in a novel NL-PE algorithm, \algname{},
which uses: 1) Natural Language Inference (NLI) between user preference utterances and NL item descriptions to maintain Bayesian preference beliefs, and 2) BO strategies such as Thompson Sampling (TS) and Upper Confidence Bound (UCB) to guide LLM query generation. We numerically evaluate our methods in controlled simulations, finding that after 10 turns of dialogue, \algname{} can achieve an MRR@10 of up to 0.27 compared to the best monolithic LLM baseline's MRR@10 of 0.17, 
despite relying on earlier and smaller LLMs.\footnote{Our code is publically available at \href{https://github.com/D3Mlab/llm-pe}{https://github.com/D3Mlab/llm-pe}.} 

\end{abstract}




\maketitle

\section{Introduction} Personalized conversational recommendation (ConvRec) systems require effective natural language (NL) preference elicitation (PE) strategies that can efficiently learn a user's top item preferences in cold start settings, ideally requiring only an arbitrary set of NL item descriptions. While the advent of large language models (LLMs) has introduced the technology to facilitate NL-PE conversations \cite{handa2024bayesian,li2023eliciting} we conjecture that monolithic LLMs 
have limited abilities to strategically conduct active, multi-turn NL-PE dialogues about a set of arbitrary items. Specifically, we hypothesize that LLMs lack the multi-turn decision-theoretic reasoning to interactively generate queries that avoid over-exploitation or over-exploration of user-item preferences, thus risking over-focusing on already revealed item preferences or wastefully exploring preferences over low-value items. Further challenges faced by monolithic LLM NL-PE approaches include the need to jointly reason over large, potentially unseen sets of item descriptions, and the lack of control and interpretability in system behaviour even after prompt engineering or fine-tuning \cite{maynez2020faithfulness}. 



\begin{figure*}[ht] 
    \centering
    \includegraphics[width = \textwidth]{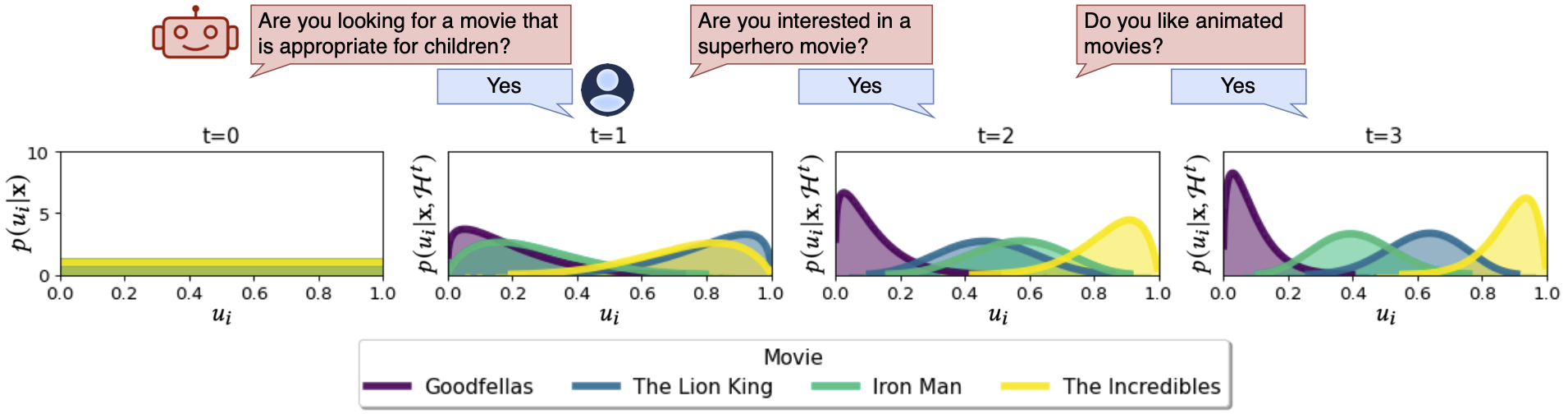}
    \caption{\algname{}'s belief updates over a cold-start user's item utilities during three turns of NL dialogue. Bayesian preference beliefs not only facilitate recommendation, but also enable Bayesian optimization policies to guide LLM query generation, avoiding over-exploration (asking about clearly low-value items) and over-exploitation (over-focusing on known preferences). 
    }
    \label{fig: posteriors}
\end{figure*}

In contrast, conventional PE algorithms \cite{li_seamless_attributes, conversational_contextual_bandit, lei_2020_path_reasoning, lee2019melu,  zhao2013interactive}, including Bayesian optimization methods \cite{guo2010real, yang2021bayesianpreference, christakopoulou2016towards, vendrov2019_evoi_pe, boutilier2002pomdp}, establish formal decision-theoretic policies such as Thompson Sampling (TS) and Upper Confidence Bound (UCB) \cite{karamanolakis2018item} to balance exploration and exploitation with the goal of quickly identifying the user's most preferred items. However, these techniques typically assume a user can express preferences via direct item ratings or comparisons -- an unrealistic expectation when users are unfamiliar with most items~\cite{biyik2023preferenceEW}. 
While recent work has extended Bayesian PE to a fixed set of template-based queries over pre-defined keyphrases~\cite{yang2021bayesianpreference}, no existing work extends Bayesian methodologies to generative NL-PE over a set of generic NL item descriptions.

In this paper, we make the following contributions:
\begin{itemize}
    \item We introduce the first Bayesian optimization formalization of NL-PE for arbitrary NL dialogue over a generic set of NL item descriptions -- establishing a new framework for research on steering LLMs with decision-theoretic reasoning.
    \item We present \algname{} (\textbf{P}reference \textbf{E}licitation with \textbf{B}ayesian \textbf{O}ptimization augmented \textbf{L}LMs), a novel NL-PE algorithm which 1) 
        infers item preferences via Natural Language Inference (NLI) \cite{yin2019benchmarking} between dialogue utterances and item descriptions to maintain Bayesian preference beliefs and 2)
        introduces LLM-based acquisition functions, where NL query generation is guided by decision-theoretic strategies such as TS and UCB over the preference beliefs. 
    \item We numerically evaluate \algname{} against monolithic GPT-3.5 and Gemini-Pro NL-PE methods via controlled NL-PE dialogue experiments over multiple NL item datasets and levels of user noise. 
    \item We observe that after 10 turns of dialogue, \algname{} can achieve a mean MRR@10 of up to 0.27 compared to the best monolithic LLM baseline's MRR@10 of 0.17, despite relying on earlier and smaller LLMs. 
\end{itemize}


\section{Background and Related Work}
\subsection{Bayesian Optimization} \label{sec: BO}
Given an objective function $f: \mathcal{X} \rightarrow \mathbb{R}$, (standard) optimization systematically searches for a point $x^* \in \mathcal{X}$ that maximizes\footnote{We take the maximization direction since this paper searches for items with maximum utility for a person.} $f$. Bayesian optimization focuses on settings where $f$ is a black-box function which does not provide gradient information and cannot be evaluated exactly -- rather, $f$ must be evaluated using indirect or noisy observations which are expensive to obtain \cite{garnett2023bayesian, 7352306}. To address these challenges, Bayesian optimization maintains probabilistic beliefs over $f(x)$ and its observations to guide an uncertainty-aware optimization policy which decides where to next observe $f(x)$. 

Bayesian optimization begins with a \textit{prior} $p(f)$ which represents the beliefs about $f$ before any observations are made. Letting $y_i$ represent a noisy or indirect observation of $f(x_i)$, and collecting a sequence of observations into a dataset $\mathcal{D} = (\mathbf{x},\mathbf{y})$, an \textit{observation model} defines the \textit{likelihood} $p(\mathcal{D}|f)$. We then use the observed data and Bayes theorem to update our beliefs and obtain the \textit{posterior} 
\begin{equation}
p(f|\mathcal{D}) = \frac{p(f)p(\mathcal{D}|f)}{p(\mathcal{D})} .
\end{equation}
This posterior informs an \textit{acquisition function} $\gamma(x|\mathcal{D})$ which determines where to next observe $f(x)$ in a way that balances exploitation (focusing observations where $f$ is likely near its maximum) with exploration (probing areas where $f$ has high uncertainty).  




\subsection{Preference Elicitation} \label{sec: bckg pe}
PE has witnessed decades of research, and includes approaches based on Bayesian optimization (e.g., \cite{eric2007active, guo2010gaussian, brochu2010bayesian, gonzalez2017preferential, bo_khajah_games}), Bandits (e.g., \cite{li2010contextual, li2011unbiased, zhao2013interactive, christakopoulou2016towards}), constrained optimization \cite{rossi2004acquiring}, and POMDPs \cite{boutilier2002pomdp}. In the standard PE setting, a user is assumed to have some hidden utilities $\mathbf{u} = [u_1,...,u_N]$ over a set $\mathcal{I}$ of $N$ items, where item $i$ is preferred to item $j$ if $u_i > u_j$. The goal of PE is typically to search for an item $i^* \in \arg\max_i u_i$ that maximizes user utility in a minimal number of PE queries, which most often ask a user to express item preferences as item ratings (e.g., \cite{li2010contextual, li2011unbiased, brochu2010bayesian, zhao2013interactive, christakopoulou2016towards}) or relative preferences between item pairs or sets (e.g., \cite{boutilier2002pomdp, guo2010real, handa2024bayesian, eric2007active, gonzalez2017preferential, vendrov2019_evoi_pe}). An alternative form of PE asks users to express preferences over predefined item features, also through rating- or comparison-based queries \cite{li_seamless_attributes, conversational_contextual_bandit, lei_2020_path_reasoning}. Central to the above PE methods are query selection strategies that balance the exploration and exploitation of user preferences, with TS and UCB algorithms (cf. Sec. \ref{sec: acquisition fun}) often exhibiting strong performance \cite{zhao2013interactive,christakopoulou2016towards,yang2021bayesianpreference, li_seamless_attributes, conversational_contextual_bandit}. However, none of these methods are able to interact with users through NL dialogue or reason about NL item descriptions.

\begin{figure*}[ht]
    \centering
    \includegraphics[width=1\linewidth]{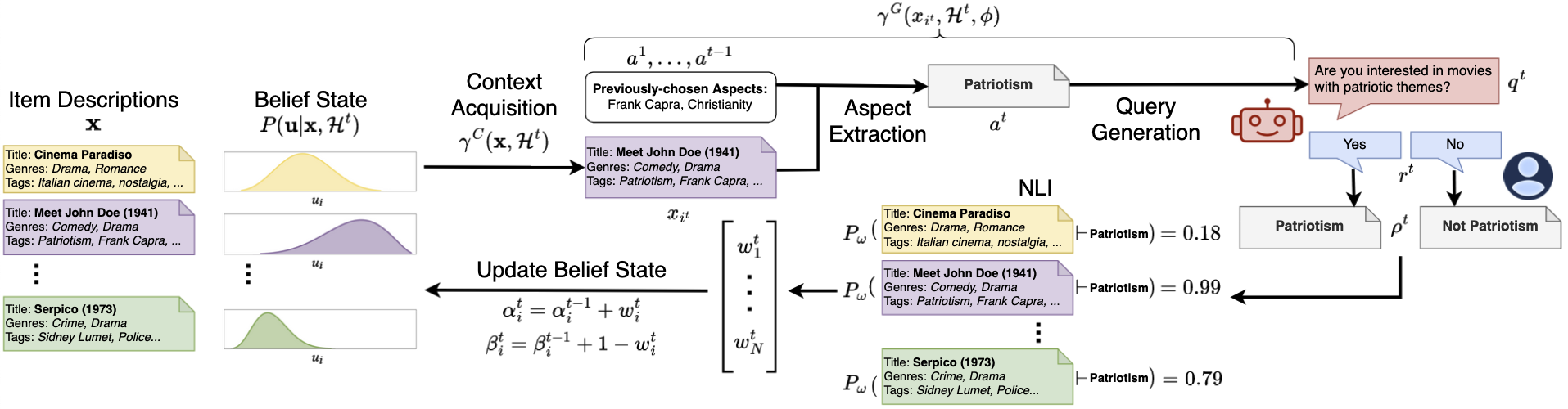}
    \caption{The \algname{} NL-PE algorithm, which maintains a Bayesian belief state over a user's item preferences given an arbitrary set of NL item descriptions $\mathbf{x}$. This belief is used by a decision-theoretic policy to balance the exploration and exploitation of preferences by strategically selecting an item description $x_{i^t}$ as the basis for LLM query generation. Belief updates are computed through Bayesian inference with NLI entailment scores between item descriptions and query-response pairs.}
    \label{fig:llmpe_diagram}
\end{figure*}




\subsection{Language-Based Preference Elicitation}
Yang \textit{et al. }\cite{yang2021bayesianpreference} introduce Bayesian PE strategies using TS and UCB for keyphrase rating queries, where keyphrases are first mined from NL item reviews and then co-embedded with user-item preferences in a recommendation system. Handa \textit{et al.} \cite{handa2024bayesian} propose using LLMs to interface with a conventional Bayesian PE system, suggesting a preprocessing step to extract features from NL descriptions and a verbalization step to fluidly express pairwise item comparison queries. Li \textit{et al.} \cite{li2023eliciting} prompt an LLM to generate PE queries for some specific domain (e.g., news content, morals), observe user responses, and evaluate LLM relevance predictions for a single item. While these works make progress towards NL-PE, they do not study how LLM query generation can strategically explore user preferences towards an arbitrary item set outside the realm of item-based or category-based feedback.

\subsection{Conversational Recommendation}
Recent work on ConvRec uses language models\footnote{Earlier systems (e.g. \cite{li2018towards,chen2019towards}) use relatively small RNN-based language models.} to facilitate NL dialogue while integrating calls to a recommender module which generates item recommendations based on user-item interaction history \cite{li2018towards, chen2019towards, wang2022towards, yang2022improving}. He \textit{et al.} \cite{he2023large} report that on common datasets, zero-shot GPT-3.5/4 outperforms these ConvRec methods, which generally use older language models and require user-item interaction history for their recommendation modules.

\subsection{Natural Language Inference} \label{sec:bckg NLI}
Binary Natural Language Inference (NLI) \cite{yin2019benchmarking} models predict the likelihood that one span of text called a premise is \textit{entailed by} (i.e., can be inferred from) a second span called the hypothesis. For example, an effective NLI model should predict a high likelihood that the premise \textit{``I want to watch Iron Man''} entails the hypothesis \textit{``I want to watch a superhero movie''}. As illustrated by this example, the hypothesis typically must be more general than the premise. NLI models are trained by fine-tuning encoder-only LLMs on NLI datasets \cite{williams2017broad,dagan2005pascal, thorne2018fever}, which typically consist of short text spans for the premise and hypothesis -- thus enabling relatively efficient performance on similar tasks with a fairly small number LLM parameters. 

\begin{figure*}[ht]
    \centering
    \includegraphics[width = 0.88
    \textwidth]{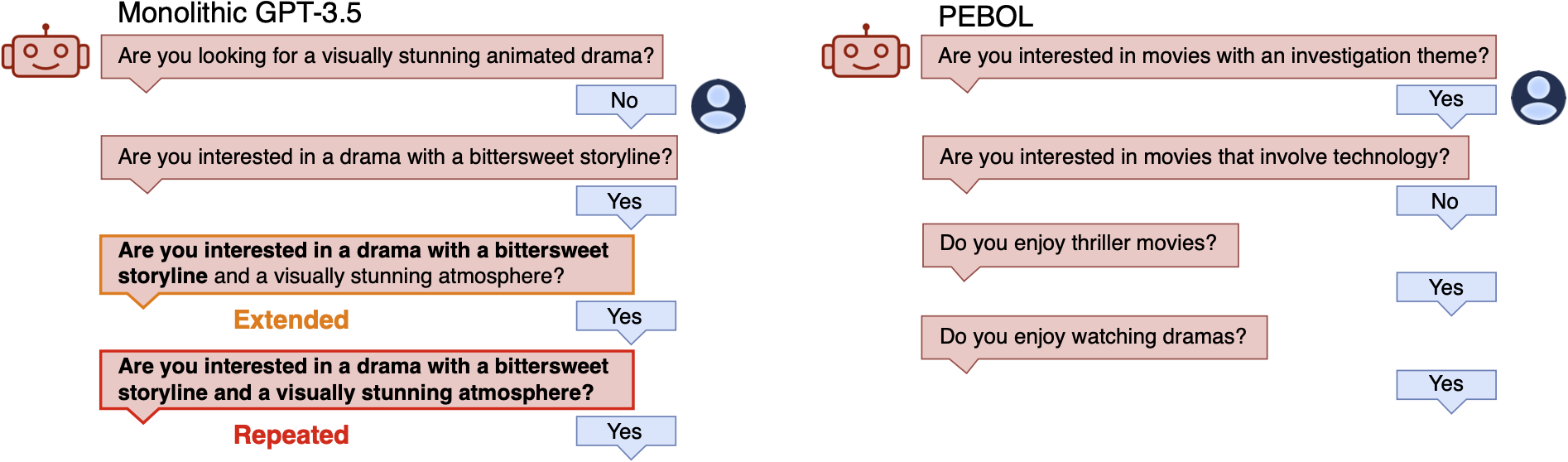}
    \caption{Cherry-picked system-generated dialogues from our NL-PE experiments. The Monolithic GPT-3.5 dialogue (left) demonstrates over-exploitation, with $q^3$ directly extending $q^2$ after a positive user preference is observed and leading to the extreme case of query repetition ($q^4$ = $q^3$). In contrast, \algname{} (right) continues exploring even after a positive response, while focusing on promising aspects (three out of four queries elicit a positive response) by using UCB-guided query generation.} 
    \label{fig:enter-label}
\end{figure*}

\section{Problem Definition} \label{sec: problem def}

We now present a Bayesian optimization formulation of NL-PE.
The goal of NL-PE is to facilitate a NL dialogue which efficiently discovers a user's most preferred items out of a set of $N$ items. Each item $i \in \mathcal{I}$ has a NL description $x_i$, which might be a title, long-form description, or even a sequence of reviews, with the item set $\mathcal{I}$ collectively represented by $\mathbf{x} \in \mathcal{X}$ with $\mathbf{x} = [x_1,...,x_N]$. We assume the user has some (unknown) utility function $f:\mathcal{X}\rightarrow\mathbb{R}$ establishing hidden utilities $\mathbf{u} = f(\mathbf{x})$ so that item $i$ is preferred to item $j$ if $u_i > u_j$. Our goal is to find the most preferred item(s): 
\begin{equation}
    i^* \in \arg \max_{i \in \mathcal{I}} u_i.
\end{equation}

In contrast to standard Bayesian PE formalisms (c.f. Sec \ref{sec: bckg pe}), we do not assume that the user can effectively convey direct \textit{item-level} preferences by either: 1) providing item ratings (i.e., utilities) or 2) pairwise or listwise item comparisons. 
Instead, we must infer user preferences by observing utterances during a NL system-user dialogue. At turn $t$ of a dialogue, we let $q^t$ and $r^t$ be the system and user utterance, respectively, with $\mathbf{q}^t = [q^1,...,q^t]$ and $\mathbf{r}^t = [r^1,...,r^t]$ representing all system and user utterances up to $t$. In this paper, we call $q^t$ the \textit{query} and $r^t$ the \textit{response}, though extensions to more generic dialogues (e.g., when users can also ask queries) are discussed in Section \ref{sec: fw}. We let $\mathcal{H}^t = (\mathbf{q}^t,\mathbf{r}^t)$ be the conversation history at turn $t$.



To formulate NL-PE as a Bayesian optimization problem, we place a prior belief on the user's utilities $p(\mathbf{u}|\mathbf{x})$, potentially conditioned on item descriptions since they are available before the dialogue begins. We then assume an observation model that gives the likelihood $p(\mathbf{r}^t|\mathbf{x},\mathbf{u},\mathbf{q}^t$), letting us define the posterior utility belief as
\begin{equation}
    p(\mathbf{u}|\mathbf{x},\mathcal{H}^{t}) \propto p(\mathbf{r}^t|\mathbf{x},\mathbf{u,\mathbf{q}^t})p(\mathbf{u}|\mathbf{x}).
\end{equation}
This posterior informs an acquisition function $\gamma(\mathbf{x},\mathcal{H}^{t})$ which generates\footnote{To represent the \textit{generative acquisition} of NL outputs, we deviate from the conventional definition of acquisition functions as mapping to $\mathbb{R}.$} a new NL query 
\begin{equation}
    q^{t+1} = \gamma(\mathbf{x},\mathcal{H}^{t}),
\end{equation}
to systematically search for $i^*$.
The preference beliefs also let us define an Expected Utility (EU) $\mu_i^t$ for every item as
\begin{equation}
    \mu_i^t = \mathbb{E}_{p(\mathbf{u}|\mathbf{x},\mathcal{H}^t)}[u_i 
    ], \label{eq: mu}
\end{equation}
which allows the top-$k$ items to be recommended at any turn based on their expected utilities. 
  
Our Bayesian optimization NL-PE paradigm lets us formalize several key questions, including:
\begin{enumerate}
    \item How do we represent beliefs $p(\mathbf{u}|\mathbf{x},\mathcal{H}^{t})$ in user-item utilities $\mathbf{u}$, given NL item descriptions $\mathbf{x}$ and a dialogue $
    \mathcal{H}^{t}$?
    \item What are effective models for the likelihood $p(\mathbf{r}^t|\mathbf{x},\mathbf{u},\mathbf{q}^t)$ of observed responses $\mathbf{r}^t$ given $\mathbf{x}$, $\mathbf{q}^t$, and user utilities $\mathbf{u}$?
    \item How can our beliefs inform the generative acquisition of NL queries $q^{t+1}$ given $\mathcal{H}^t$ to strategically search for $i^*$?
\end{enumerate}
These questions reveal a number of novel research directions discussed further in Section \ref{sec: fw}. In this paper, we present \algname{}, a NL-PE algorithm based on the above Bayesian optimization NL-PE formalism, and numerically evaluate it against monolithic LLM alternatives through controlled, simulated NL dialogues (cf. Sec. \ref{sec: experimental results}). 



\section{Methodology} \label{sec:meth}

\paragraph{Limitations of Monolithic LLM Prompting} An obvious NL-PE approach, described further as baseline in Section \ref{sec: monollm explained}, is to prompt a monolithic LLM with all item descriptions $\mathbf{x}$, dialogue history $\mathcal{H}^{t}$, and instructions to generate a new query at each turn. However, providing all item descriptions $[x_1,...,x_N]$ in the LLM context window is very computationally expensive for all but the smallest item sets. While item knowledge could be internalized through fine-tuning, each item update would imply system retraining. Critically, an LLM's preference elicitation behaviour cannot be controlled other than by prompt-engineering or further fine-tuning, with neither option offering any guarantees of predictable or interpretable behaviour that balances the exploitation and exploration of user preferences.   

\paragraph{\algname{} Overview} We propose to addresses these limitations by augmenting LLM reasoning with a Bayesian Optimization procedure in a novel algorithm, \algname, illustrated in Figure \ref{fig:llmpe_diagram}. 
At each turn $t$, our algorithm maintains a probabilistic belief state over user preferences as a Beta belief state 
(cf. Sec. \ref{sec:utility beliefs}).
This belief state guides an LLM-based acquisition function to generate NL queries explicitly balancing exploration and exploitation to uncover the top user preferences (cf. Sec. \ref{sec: acquisition fun}). In addition, our acquisition function reduces the context needed to prompt the LLM in each turn from all $N$ item descriptions $\mathbf{x}$ to a single strategically selected item description $x_{i^t}$. \algname{} then uses NLI over elicited NL preferences and item descriptions to map dialogue utterances to numerical observations (c.f. Sec \ref{sec: nli}). 

\begin{figure*}[ht]
    \centering
    \includegraphics[width=1\linewidth]{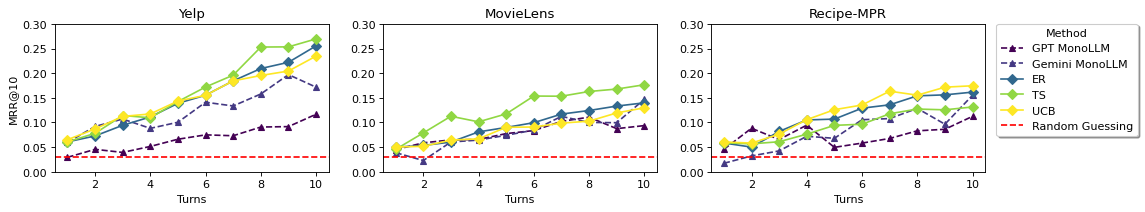}
    \caption{MRR@10 for MonoLLM and \algname{-P} with uncertainty-informed policies (UCB, TS, ER). All methods show preference learning over time and MonoLLM is generally outperformed by \algname{}. }
    \label{fig:noise0_map_plot}
\end{figure*}

\subsection{Utility Beliefs} \label{sec:utility beliefs}
\subsubsection{Prior Beliefs} \label{sec:prior} Before any dialogue, \algname{} establishes an uninformed prior belief $p(\mathbf{u})$ on user-item utilities.  We assume item utilities are independent so that \begin{equation}
    p(\mathbf{u}) = \prod_{i = 1}^N p(u_i), \label{eq:util_indep}
\end{equation}  
and that the prior for each utility $u_i$ is a Beta distribution \begin{equation}
    p(u_i) = \text{Beta}(\alpha_i^0,\beta_i^0). \label{eq:prior}
\end{equation} 
Since this paper focuses on fully cold start settings, we assume a uniform Beta prior with $(\alpha_i^0,\beta_i^0) = (1,1)$. Beta distributions, illustrated in Figure \ref{fig: posteriors}, lie in the domain $[0,1]$ -- a normalized interval for bounded ratings in classical recommendation systems. We can thus interpret utility values of $u_i = 1$ or $u_i = 0$ to represent a complete like or dislike of item $i$, respectively, while values $u_i \in (0,1)$ provide a strength of preference between these two extremes.

\subsubsection{Observation Model} \label{sec:likelihood}
To perform a posterior update on our utility beliefs given observed responses $\mathbf{r}^t$, we need an observation model that represents the likelihood $p(\mathbf{r}^t|\mathbf{x},\mathbf{u},\mathbf{q}^t)$. Modelling the likelihood of $\mathbf{r}^t$ is a challenging task, so we will require some simplifying assumptions. Firstly, we assume that the likelihood of a single response $r^t$ is independent from any previous dialogue history $\mathcal{H}^{t-1}$, so that: 
\begin{equation}
    p(\mathbf{r}^t|\mathbf{x},\mathbf{u},\mathbf{q}^t) = \prod_{t'=1}^t p(r^{t'}|\mathbf{x},\mathbf{u}, q^{t'}).
\end{equation}
Note that this independence assumption will allow incremental posterior belief updates, so that 
\begin{equation}
    p(\mathbf{u}|\mathbf{x},\mathcal{H}^t) \propto p(r^{t}|\mathbf{x},\mathbf{u},q^{t})
    p(\mathbf{u}|\mathbf{x},\mathcal{H}^{t-1}). \label{eq:incremental updates}
\end{equation}

\subsubsection{Binary Item Response Likelihoods and Posterior Update} 
With the factorized distributions over item utilities and observational likelihood history now defined, we simply have to provide a concrete observational model of the response likelihood conditioned on the query, item descriptions, and latent utility:  $p(r^{t}|\mathbf{x},\mathbf{u},q^{t})$.

Because the prior is factorized over conditionally independent $u_i$ (cf.~\eqref{eq:util_indep}), we can likewise introduce individual per-item factorized binary responses $r^t_i \in \{ 0 (\mathit{dislike}), 1 (\mathit{like}) \}$ to represent the individual relevance of each item $i$ to the preference elicited at turn $t$.  Critically, we won't actually require an individual response per item --- this will be computed by a natural language inference (NLI) model~\cite{dagan2005pascal} to be discussed shortly --- but we'll begin with an individual binary response model for $r_i^t$ for simplicity:
\begin{equation}
    p(r_i^t|x_i,u_i,q^{t}) = \mbox{Bernoulli}(u_i).  \label{eq:likelihood}
\end{equation}
%

With our response likelihood defined, this now leads us to our first pass at a full posterior utility update that we term \algname{-B} for observed \emph{Binary} rating feedback.  Specifically, given observed binary ratings $r^t_i$, the update at $t=1$ uses the Beta prior \eqref{eq:prior} with the Bernoulli likelihood \eqref{eq:likelihood} to form a standard Beta-Bernoulli conjugate pair and compute the posterior utility belief
\begin{align}
    p(u_i|x_i,\mathcal{H}^1) &\propto p(u_i|x_i) p(r^1_i|x_i,u_i,q^{t})\\
                              &= \text{Beta}(\alpha_i^1,\beta_i^1),
\end{align}
where $\alpha_i^1 = \alpha_i^0 + r_i^1$, $\beta_i^1 = \beta_i^0 + (1-r_i^1)$. Subsequent incremental updates updates follow Eq. \eqref{eq:incremental updates} and use the same conjugacy to give 
\begin{align}
    p(u_i|x_i,\mathcal{H}^t) = \text{Beta}(\alpha_i^t,\beta_i^t), \label{eq:binary posterior}
\end{align}
where $\alpha_i^t = \alpha_i^{t-1} + r_i^t$, $\beta_i^t = \beta_i^{t-1} + (1-r_i^t)$.

\subsubsection{Natural Language Inference and Probabilistic Posterior Update} \label{sec:post inf}
As hinted above, effective inference becomes slightly more nuanced since we don't need to observe an explicit binary response \emph{per item} in our \algname{} framework.  Rather, we receive general preference feedback $r^t$ on whether a user generically prefers a text description $q^t$ and then leverage an NLI model~\cite{dagan2005pascal} to \emph{infer} whether the description $x_i$ of item $i$ would be preferred according to this feedback.
For instance, for a $(q^t,r^t)$ pair (\textit{``Want to watch a children's movie?'',``Yes}''), NLI should infer a rating of $r_1^t = 1$ for $x_1 = $ \textit{``The Lion King''} and $r_2^t = 0$ for $x_2 = $ \textit{``Titanic''}. 

To deal with the fact that NLI models actually return an \emph{entailment probability}, our \textit{probabilistic} observation variant, \algname{-P} leverages the probability that 
item description $x_i$ entails $q_t$, 
which we denote as $w_i^t$. 
We provide a full graphical model and derivation of the Bayesian posterior update given this entailment probability in the Supplementary Material, but note that we can summarize the final result as a \emph{relaxed} version of the binary posterior update of~\eqref{eq:binary posterior} that replaces the binary observation $r_i \in \{ 0,1 \}$ with the entailment probability $w_i^t \in [0,1]$, i.e., 
$\alpha_i^t = \alpha_i^{t-1} + w_i^t$, $\beta_i^t = \beta_i^{t-1} + (1-w_i^t)$.

To visually illustrate how this posterior inference process works in practice, Figure \ref{fig: posteriors} shows the effect of \algname{'s} posterior utility belief updates based on NLI for three query-response pairs -- we can see the system gaining statistical knowledge about useful items for the user from the dialogue. 





\begin{figure*}[ht]
    \includegraphics[width=1\linewidth]{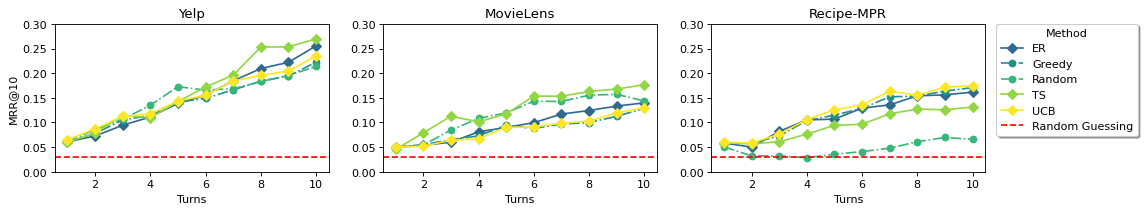}
    \caption{MRR@10 for \algname{-P} with various context acquisition policies.}
    \label{fig:noise0_map_plot_others}
\end{figure*}

\subsection{LLM-Based Acquisition Functions} \label{sec: acquisition fun}
Recall from Sec. \ref{sec: BO} that in Bayesian optimization, the posterior informs an acquisition function which determines where to make the next observation.
\algname{} generates a new query $q^t$ with a two-step acquisition function $\gamma$, first using Bayesian Optimization policies (step 1) based on the posterior utility beliefs $p(\mathbf{u}|\mathbf{x},\mathcal{H}^{t})$ to select NL context, and then using this selected context to guide LLM prompting (step 2). 
We express the overall acquisition function $\gamma = \gamma^G \circ \gamma^C$ as a composition of a \textit{context} acquisition function $\gamma^C$ (cf. Sec. \ref{sec:context acq}) and a NL \textit{generation} function $\gamma^G$ (cf. Sec. \ref{sec:lang gen}). 

\subsubsection{Context Acquisition via Bayesian Optimization Policies} \label{sec:context acq}

First, \algname{} harnesses Bayesian optimization policies to select an item description $x_{i^t}$ which will be used to prompt an LLM to generate a query about an aspect described by $x_{i^t}$ (cf. Sec. \ref{sec:lang gen}). Selecting an item $i^t$ whose utility $u_{i^t}$ is expected to be near the maximum, $u_{i^*}$, will generate \textit{exploitation queries} asking about properties of items that are likely to be preferred by the user. In contrast, selecting an item $i^t$ associated with high uncertainty in its utility $u_i^t$ will generate \textit{exploration queries} that probe into properties of items for which user preferences are less known. Thus, strategically selecting $x_{i^t}$ allows \algname{} to balance the exploration and exploitation behaviour of NL queries, decreasing the risks of becoming stuck in local optima (over-exploitation) or wasting resources exploring low utility item preferences (over-exploration). We define the item selected by the context acquisition function as
\begin{equation}
    i^t = \gamma^C(\mathbf{x},\mathcal{H}^{t}),
\end{equation} and list several alternatives for $\gamma^C$, including the well-known strategies of TS and UCB \cite{7352306}: 
\begin{enumerate}
    \item \textbf{Thompson Sampling (TS)}: First, a sample of each item's utility $\hat{u}^t_i$ is taken from the posterior, $\hat{u}^t_i \sim p(u_i|x_i,\mathcal{H}^{t}).$ Then, the item with the highest sampled utility is selected:
    \begin{equation}
        i^t = \arg\max_i \hat{u}^t_i.
    \end{equation}
    TS explores more when beliefs have higher uncertainty and exploits more as the system becomes more confident.
    \item \textbf{Upper Confidence Bound (UCB)}: Let $P_k(\alpha,\beta)$ represent the $k$'th percentile of $\text{Beta}(\alpha,\beta)$, which provides a confidence bound on the posterior. UCB selects the item with the highest confidence bound
        \begin{equation}
        i^t = \arg\max_i P_k(p(u_i|x_i,\mathcal{H}^{t})),
    \end{equation}
    following a balanced strategy because confidence bounds are increased by both high utility and high uncertainty. 
    \item \textbf{Entropy Reduction (ER)}: An explore-only strategy that selects the item with the most uncertain utility:
    \begin{equation}
        i^t = \arg\max_i \text{Var}(p(u_i|x_i,\mathcal{H}^{t})).
    \end{equation}
    \item \textbf{Greedy}: An exploit-only strategy that selects the item with the highest expected utility $\mu_i^t$ (Eq. \ref{eq: mu}): 
    \begin{equation}
        i^t = \arg\max_i \mu_i^t.
    \end{equation}
    \item \textbf{Random}: An explore-only heuristic that selects the next item randomly.
\end{enumerate}










\subsubsection{Generating Short, Aspect-Based NL Queries} \label{sec:lang gen}
Next, \algname{} prompts an LLM to generate a NL query $q^t$ based on the selected item description $x_{i^t}$ while also using the dialogue history $\mathcal{H}^{t}$ to avoid repetitive queries. We choose to generate ``yes-or-no'' queries asking if a user prefers items with some aspect $a^t$, which is a short text span extracted dynamically from $x_{i^t}$ to be different from any previously queried aspects $a^1,...,a^{t-1}$. We adopt this query generation strategy to: 1) reduce cognitive load on the user, who may be frustrated by long and specific queries about unfamiliar items and 2) better facilitate NLI through brief, general phrases \cite{yin2019benchmarking}. Letting $\phi$ represent the query generation prompt, we let
\begin{equation}
    q^t, a^t = \gamma^G(x_{i^t}, \mathcal{H}^{t},\phi)
\end{equation}
be the LLM generated query and aspect at turn $t$, with prompting details discussed in Section \ref{sec:query gen implementation}. An example of such a query and aspect (bold) is \textit{``Are you interested in movies with \textbf{patriotic} themes?''}, generated by \algname{} in our movie recommendation experiments and shown in Figure \ref{fig:llmpe_diagram}.

\subsection{NL Item-Preference Entailment} \label{sec: nli}
\subsubsection{Preference Descriptions from Query Response Pairs} 
Next, \algname{} receives a NL user response $r^t$, which it must convert to individual item preference observations.
Since the LLM is instructed  to generate "yes-or-no" queries $q^t$ asking a user if they like aspect $a^t$, 
we assume the user response will be a \textit{"yes"} or a \textit{"no"}, and 
create a NL description of the users preference $\rho^t$, letting $\rho^t = a^t$ if $r^t=$\textit{``yes''}, and $\rho^t = \text{concat(\textit{``not ''}},a^t)$ if $r^t=$ \textit{``no''}. For example, given a query that asks if the user prefers the aspect ``patriotism'' in an item, if the user response is \textit{``yes''}, then the user preference $\rho^t$ is ``patriotism'', and ``not patriotism'' otherwise. 
This approach produces short, general preference descriptions that are well suited for NLI models \cite{yin2019benchmarking}.

\subsubsection{Inferring Item Ratings from NL Preferences} \label{NLI entailment}
Given a NL preference $\rho^t$, \algname{} must infer whether the user would like an item described by $x_i$. 
Specifically, \algname{} acquires ratings $\mathbf{w}^t = [w^t_1,...,w^t_N]$ (cf. Sec. \ref{sec:post inf}) by using NLI to predict whether an item description $x_i$ entails (i.e., implies) the preference $\rho^t$. For example, we expect that an NLI model would predict that $x_i =$``\textit{The Lion King}'' entails $\rho^t=$\textit{\textbf{``animated''}} while $x_j=$\textit{``Titanic''} does not, inferring that a user who expressed preference $\rho^t$ would like item $i$ but not $j$. We use an NLI model $P_{\omega}(x_i,\rho^t)$ to predict the probability $w_i^t$ that $x_i$ entails $\rho^t$, and return 
$r_i^t = \lfloor w_i^t \rceil$
in the case of binary observations (\algname{-B}) and $w_i^t$ in the case of probabilistic observations (\algname{-P}). 

\subsection{The Complete \algname{} System}

This concludes the \algname{} specification -- the entire process from prior utility belief to the LLM-based acquisition function generation of a query to the posterior utility update is illustrated in Figure \ref{fig:llmpe_diagram}.

\begin{figure*}[ht]
    \centering
    \includegraphics[width=1\linewidth]{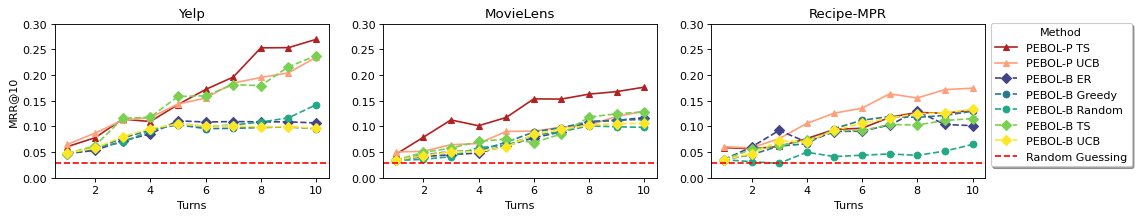}
    \caption{MRR@10 for \algname{} using binary vs. probabilistic entailment scores. \algname{-P} with the best policy (TS on Yelp and MovieLens, UCB on Recipe-MPR) generally outperforms \algname{-B}.}
    \label{fig:binary_vs_continuous}
\end{figure*}


\section{Experimental Methods} \label{sec: experimental methods}
We numerically evaluate our \algname{} variations through controlled NL-PE dialogue experiments across multiple datasets and response noise levels -- comparing against two monolithic LLM (MonoLLM) baselines. Specifically, these baselines directly use  GPT-3.5-turbo-0613 (GPT MonoLLM) or Gemini-Pro (Gemini MonoLLM) as the NL-PE system, as described in Section \ref{sec: monollm explained}. 
We do not compare against ConvRec methods \cite{li2018towards, chen2019towards, wang2022towards, yang2022improving} because they are not cold-start systems, requiring observed user-item interactions data to drive their recommendation modules. We also do not base our experiments on ConvRec datasets such as ReDIAL \cite{li2018towards}, since they are made up of \textit{pre-recorded} conversation histories and cannot be used to evaluate active, cold-start NL-PE systems.

\subsection {MonoLLM Baseline} \label{sec: monollm explained}
A major challenge of using MonoLLM for NL-PE is that item descriptions $\mathbf{x}$ either need to be internalized through training or be provided in the context window (cf. Sec. \ref{sec:meth}) -- since we focus on fully cold-start settings, we test the latter approach as a baseline. In each turn, given the full conversation history $\mathcal{H}^t$ and $\mathbf{x}$, we prompt the MonoLLM to generate a new query to elicit user preferences  -- all prompts are shown in the Supplementary Materials. We evaluate recommendation performance after each turn by using another prompt 
to recommend a list of ten item names from $\mathbf{x}$ given $\mathcal{H}^t$. Due to context window limits, this MonoLLM approach is only feasible for small item sets with short item descriptions; thus, we have to limit $|\mathcal{I}|$ to 100 for fair comparison to the MonoLLM baseline.


\subsection{Simulation Details}
We test \algname{} and MonoLLM through NL-PE dialogues with LLM-simulated users, where the simulated users' item preferences remain hidden from the system. We evaluate recommendation performance over 10 turns of dialogue. 

\subsubsection{User Simulation} \label{user simulation}
For each experiment, we simulate 100 users, each of which likes a single item $i \in \mathcal{I}$. Each user is simulated by GPT-3.5-turbo-0613, which is given item description $x_i$ and instructed to provide only ``yes'' or ``no'' responses to a query $q^t$ as if it was a user who likes item $i$. 

\subsubsection{Evaluating Recommendations}
We evaluate the top-10 recommendations in each turn using the Mean Reciprocal Rank (MRR@10) of the preferred item, which is equivalent to MAP@10 for the case of a single preferred item. 

\subsubsection{\algname{} Query Generation} \label{sec:query gen implementation}
In turn $t$, given an item description $x_i$ and previously generated aspects $(a^1,...,a^{t-1})$, an LLM (GPT-3.5-turbo-0613)\footnote{Experiments with newer LLMs such as Gemini or GPT4 for PEBOL query generation are left for future work due to the API time and cost requirements needed to simulate the many variants of PEBOL reported in Section \ref{sec: experimental results}. We do, however, compare PEBOL against a Gemini-MonoLLM baseline.} is prompted to generate an aspect $a^t$ describing the item $i$ that is no more than 3 words long. 
The LLM is then prompted again to generate a ``yes-or-no'' query asking if a user prefers $a^t$. 

\subsubsection{NLI} \label{entailment}
We use the 400M FAIR mNLI\footnote{\href{https://huggingface.co/facebook/bart-large-mnli}{https://huggingface.co/facebook/bart-large-mnli}} model to predicts logits for \textit{entailment}, \textit{contradiction}, and \textit{neutral}, and divide these logits by an MNLI temperature $T \in \{1,10,100\}$  
As per the FAIR guidelines, we pass the temperature-scaled \textit{entailment} and \textit{contradiction} scores through a softmax layer and take the \textit{entailment} probabilities. We report \algname{} results using the best MNLI temperature for the most datasets.

\subsubsection{User Response Noise} We test three user response noise levels $\in \ $\{0,0.25,0.5\} corresponding to the proportion or user responses that are randomly selected between "yes" and "no". %

\subsubsection{Omitting Query History Ablation} We test how tracking query history in \algname{} effects performance with an ablation study that removes previously generated aspects $(a^1,...,a^{t-1})$ from the aspect extraction prompt.


\subsection{Datasets}
We obtain item descriptions from three real-world datasets: MovieLens25M\footnote{https://grouplens.org/datasets/movielens/25m/}, Yelp\footnote{https://www.yelp.com/dataset}, and Recipe-MPR \cite{recipe_mpr} (example item descriptions from each shown in Table 1 in the Supplementary Materials). After the 
filtering steps below for Yelp and MovieLens, we randomly sample 100 items to create $\mathbf{x}$. For Yelp, we filter restaurant descriptions to be from a single major North American city (Philadelphia) and to have at least 50 reviews and five or more category labels.
For MovieLens,\footnote{For all experiments with MovieLens, we use the 16k version of GPT-3.5-turbo-0613, due to MonoLLM requiring extra context length for $\mathbf{x}.$} we filter movies to be in the 10\% by rating count with at least 20 tags, and let movie descriptions use the title, genre labels, and 20 most common user-assigned tags. 

\begin{figure*}[ht]
    \centering
    \includegraphics[width=1\linewidth]{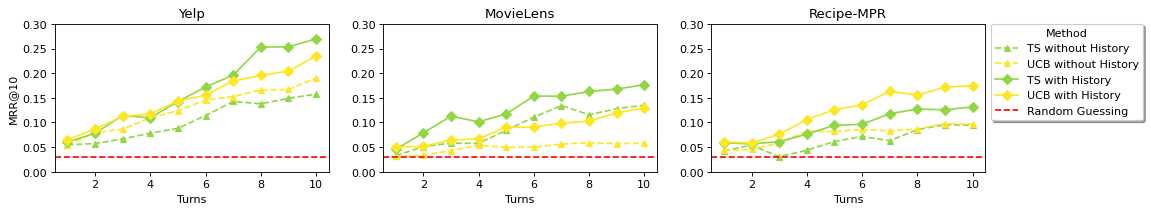}
    \caption{The effect of including the generated aspect history in the aspect generation prompt. Including the history improves performance, which we hypothesize is due to reducing repeated or uninformative queries.}
    \label{fig:history}
\end{figure*}

\begin{figure*}[ht]
    \centering
    \includegraphics[width=1\linewidth]{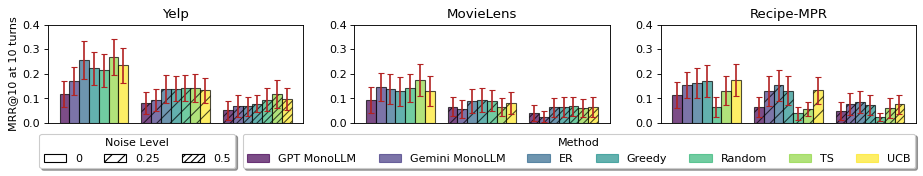}
    \caption{The effect of user response noise on MRR@10 -- error bars are 95\% confidence intervals.}
    \label{fig:noise_results}
\end{figure*}


\subsection{Research Questions}
Our experiments explore the following research questions (RQs):
\begin{itemize}
    \item \textbf{RQ1:} How does \algname{} perform against the MonoLLM baselines?
    \item \textbf{RQ2:} Does \algname{} perform better with binary or probabilistic observations, and how sensitive is the latter to temperature? 
    \item \textbf{RQ3:} How do \algname{} and MonoLLM  perform under user response noise?
    \item \textbf{RQ4:} How do the context selection policies of TS, UCB, ER, Greedy, and Random effect \algname{} performance?
    \item \textbf{RQ5:} How much does \algname{} performance depend on access to the query history during query generation?
\end{itemize}



\section{Experimental Results} \label{sec: experimental results}

\subsection{RQ1 - \algname{} vs. MonoLLM} \label{sec: Bandit vs LLM}
Figure \ref{fig:noise0_map_plot} shows MRR@10 over 10 dialogue turns for MonoLLM and PEBOL (UCB,TS,ER),\footnote{For each PEBOL policy, we use the MNLI temperature that performed best on the most datasets with continuous responses (see Supplementary Materials).} with 95\% confidence intervals (CIs) at turn 10 shown in Figure \ref{fig:noise_results} (see Supplementary Materials for CIs for all turns and experiments). All methods start near random guessing, reflecting a cold start, and show clear preference learning over time. 

Compared to GPT-MonoLLM, which uses the same LLM (GPT-3.5-turbo-0613) as PEBOL does for query generation, after 10 turns of dialogue PEBOL achieves: a mean MRR@10 of 0.27 vs. GPT-MonoLLM's MRR@10 of 0.12 on Yelp; 0.18 vs 0.09 on MovieLens; and 0.17 vs 0.11 on Recipe-MPR, respectively. Compared to Gemini-MonoLLM, which uses a newer generation LLM (Gemini-Pro) than PEBOL for query generation, after 10 turns PEBOL still achieves a higher mean MRR@10 of 0.27 vs. Gemini-MonoLLM's mean MRR@10 of 0.17 on Yelp; 0.18 vs. 0.15 on MovieLens, and 0.17 vs 0.16 on Recipe-MPR, respectively. While we did not have the resources to test PEBOL with Gemini query generation, we hypothesize that using a newer LLM for query generation can further improve PEBOL performance, since using the newer LLM (Gemini) shows performance improvements for the MonoLLM baseline.



\subsection{RQ2 - Binary vs. Probabilistic Responses} \label{sec: binary vs continuous}
Figure \ref{fig:binary_vs_continuous} compares \algname{} performance using binary (\algname{-B}) vs. continuous (\algname{-P}) feedback, and shows that performance is typically better when continuous responses are used -- indicating that binary feedback models discard valuable information from the entailment probabilities.



\subsection{RQ3 - Effect of User Response Noise} \label{sec: effect of noise}
Figure \ref{fig:noise_results} shows the impact of user response noise on MRR@10 at turn 10 -- \algname{} generally continues to outperform MonoLLM under user response noise. Specifically, at all noise levels, both MonoLLM baselines are outperformed  by all \algname{-P} variants on Yelp, and by at least one \algname{-P} variant on MovieLens and Recipe-MPR. 



\subsection{RQ4 - Comparison of Context Acquisition Policies} \label{sec: bandits strategy results}

Figure \ref{fig:noise0_map_plot_others} compares the performance of various \algname{} context acquisition policies -- all policies show active preference learning, other than random item selection on RecipeMPR. There is considerable overlap between methods, however for most turns TS does well on Yelp and MovieLens while being beaten by Greedy, ER, and UCB on Recipe-MPR. As expected due to the randomness in sampling, TS performance is correlated with random item selection, while UCB performs quite similarly to greedy.

\subsection{RQ5 - Effect of Aspect History in Query Generation} \label{sec: response history results}
As shown in Figure \ref{fig:history}, we see improvements in \algname{} performance from including a list of previously generated aspects in the aspect generation prompt. 
For instance, the differences in mean MRR@10 from including vs. excluding the query history for TS after 10 turns  were: 0.27 vs 0.16 for Yelp; 0.18 vs 0.14 for MovieLens, and 0.13 vs 0.09 for Recipe-MPR, respectively. 
Practically, including the aspect generation history also helps to avoid repeat queries, which gain no information and could frustrate a user.


\section{Conclusion and Future Work} \label{sec: fw}
This paper presents a novel Bayesian optimization formalization of natural language (NL) preference elicitation (PE) over arbitrary NL item descriptions, as well as introducing and evaluating \algname{}, an algorithm for NL \textbf{P}reference \textbf{E}licitation with \textbf{B}ayesian \textbf{O}ptimization augmented \textbf{L}LMs. As discussed below, our study also presents many opportunities for future work, including for addressing some of the limitations of \algname{} and our experiment setup.

\paragraph{User Studies} Firstly, our experiments limited by their reliance on LLM-simulated users. While the dialogue simulations indicate reasonable behaviour in the observed results, such as initial recommendation performance near random guessing, preference learning over time, and coherent user responses in logs such as those shown in Figure \ref{fig:history}, future work would benefit from human user studies. 

\paragraph{Multi-Item Belief Updates} While the assumption that item utilities can be updated independently allows the use of a simple and interpertable Beta-Bernouilli update model for each item, it also requires a separate NLI calculation to be performed for each item, which is computationally expensive. A key future direction is thus to explore alternative belief state forms which enable the joint updating of beliefs over all items from a single NLI computation. 

\paragraph{Collaborative Belief Updates} Since \algname{} does not leverage any historical interactions with other users, an important future direction is to study NL-PE which leverages collaborative, multi-user data. One possibility is to initialize a cold start user's prior beliefs based on interaction histories with other users. Another direction is adapting collaborative filtering based belief updating, such as the methods used in item-based feedback PE techniques (e.g., \cite{christakopoulou2016towards}), to NL-PE.  

\paragraph{Diverse Query Forms}
While \algname{} uses a pointwise query generation strategy that selects one item description at a time for LLM context, future work can explore LLM-based acquisition functions with pairwise and setwise context selection. Such multi-item context selection would enable \textit{contrastive} query generation that could better \textit{discriminate} between item preferences. 

\paragraph{NL-PE in ConvRec Architectures} Another direction for future research is the integration of NL-PE methodologies such as \algname{} into conversational recommendation (ConvRec) system architectures (e.g., \cite{friedman2023leveraging,kemper2024retrieval,deldjoo2024review, korikov2024llm}), which must balance many tasks including recommendation, explanation, and personalized question answering. Thus, in contrast to \algname{'s} pointwise queries and \textit{``yes-or-no''} user responses, the use of PE in ConvRec systems implies that future algorithms will need to elicit preferences based on \textit{arbitrary} pairs of NL system-user utterances. In these potential extensions, aspect-based NLI could be enabled by extracting aspects from utterances with LLMs \cite{korikov2024multi}.  

\bibliographystyle{ACM-Reference-Format}
\bibliography{refs}

\appendix

\section{Probabilistic Graphical Model for Posterior Utility Update}

\begin{figure} [t]
    \centering
    \includegraphics[width=0.6\linewidth]{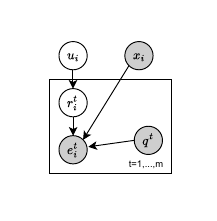}
    \caption{Graphical model used for posterior utility updates. $u_i$ is the random variable representing the utility of item $i \in \mathcal{I}$ and $x_i$ is the item description. $q^t$ is the query presented at iteration $t$, $r_{i}^{t}$ is the variable representing the latent (not directly observed) relevance of item $i$ at step $t$, and $e_{i}^{t}$ is the binary observation representing whether the item description entails the user preference. Unshaded and shaded nodes indicate unobserved and observed variables respectively.}
    \label{fig:pgm}
\end{figure}

In this section, we present the probabilistic graphical model for the posterior utility updates introduced in our paper with a more detailed derivation of the posterior utility belief. As discussed in the paper, the objective of posterior inference is to update the prior belief maintained over the utility of each item $i \in \mathcal{I}$ denoted by $p(u_i)$ given the query $q^{t}$, item description $x_i$, and $e^{t}_{i}$, the binary observation variable representing whether the item description entails the user's response to the queried preference $q_t$.

Figure \ref{fig:pgm} shows the graphical model representation of these variables. The presented query $q^{t}$ and the item description $x_i$ are observed (shaded), while the relevance of item $i$ to query $q^{t}$ is latent (unshaded) and denoted by $r_{i}^{t} \in \{0,1\}$ and conditioned on the latent item utility $u_i$.  
We observe whether the item $x_i$ ``truly'' entails (i.e.,~$e_{i}^{t} =\text{True}$) the user's response to query $q_t$ (as determined by the NLI entailment probability $w^{t}_{i}$ if the item is relevant, i.e., $r_{i}^{t}=1$).
The conditional probability distributions in this graphical model are formally defined as follows:
\begin{align}
    p(u_i) & = \text{Beta}(u_i;\alpha_i, \beta_i),
    \label{prior} \\
    p(r_{i}^{t}|u_i) & = \text{Bernoulli}(u_i),
    \label{relevance} \\
    p(e_{i}^{t} =\text{True}|r_{i}^{t}, q^{t}, x_i) & = \begin{cases}
        w^{t}_{i} &  r_{i}^{t} = 1\\
        1-w^{t}_{i} & r_{i}^{t} = 0
    \end{cases},
    \label{entailment}
\end{align}
To further explain the rationale for  Eq~\eqref{entailment}, we note that
$w^{t}_{i}$ is 
the natural language entailment probability 
that item description $x_i$ entails the aspect queried in the user's response to $q_t$ 
given that item $i$ is relevant ($r_{i}^{t}=1$).  This entailment probability is obtained from the NLI model, which produces the probability that the entailment is \emph{true}, hence the reason why $e^{t}_{i} =\text{True}$.
If item $i$ is instead irrelevant ($r_{i}^{t}=0$) then we assume for simplicity that $1-w_i^t$ is the probability of the true entailment.\footnote{We note that an alternative approach (not used here) could attempt to use NLI to determine the probability of an incorrect true entailment (or confusion) given that item $i$ is irrelevant ($r_{i}^{t}=0$).  That is, there is no inherent requirement for the two cases of Eq~\eqref{entailment} to sum to 1 since $r_{i}^{t}$ is on the conditional side.}

To obtain the posterior utility $p(u_i|x_i, q^{t}, e_{i}^{t})$, we need to marginalize the joint distribution $p(u_i,r_{i}^{t}|x_i, q^{t}, e_{i}^{t})$ over $r_{i}^{t}$. Formally, 
\begin{equation}
    p(u_i|x_i, q^{t}, e_{i}^{t}) = \sum_{r_{i}^{t}} p(u_i,r_{i}^{t}|x_i, q^{t}, e_{i}^{t}).
    \label{eq:pgm1}
\end{equation}
Considering the conditional independencies determined from the graphical model, the joint distribution factorizes as
\begin{equation}
p(u_i,r_{i}^{t}|x_i, q^{t}, e_{i}^{t}) = p(u_i)p(r_{i}^{t}|u_i)p(e_{i}^{t}|r_{i}^{t}, x_i, q^{t}).
    \label{eq:pgm2}
\end{equation}
Next, we replace the probability distribution of each factor according to Equations \eqref{prior}, \eqref{relevance}, \eqref{entailment} in  \eqref{eq:pgm1}, to obtain
\begin{multline}
   p(u_i|x_i, q^{t}, e_{i}^{t}=\text{True}) \\\propto \sum_{r_{i}^{t}\in \{0,1 \}} u_{i}^{\alpha}(1-u_i)^{\beta}\left(\begin{cases}
       u_i & r_{i}^{t} = 1\\
       1-u_i & r_{i}^{t}=0
   \end{cases}\right)\left(\begin{cases}
w^{t}_{i}& r_{i}^{t} = 1\\
       1-w^{t}_{i}& r_{i}^{t} = 0
   \end{cases}\right).
\end{multline}
Expanding the summation yields
\begin{multline}
    p(u_i|x_i, q^{t}, e_{i}^{t}=\text{True}) \propto w^{t}_{i} u_i^{\alpha+1}(1-u_i)^{\beta}+(1-w^{t}_{i})u_{i}^{\alpha}(1-u_i)^{\beta+1}\\
    \propto w^{t}_{i} \text{Beta}(u_i;\alpha+1, \beta)+(1-w^{t}_{i}) \text{Beta}(u_i;\alpha, \beta+1)
\label{mixture}
\end{multline}
The latter term represents a mixture of Beta distributions that is challenging to handle since multiple posterior updates 
would cause the number of components in the mixture to grow exponentially with the number of query observations $m$, leading to substantial computational and memory complexity. 

To address this issue, several methods have been proposed for approximating the posterior distribution to allow for tractable computations. In this work, we use the Assumed Density Filtering (ADF) approach, a technique widely used in Bayesian filtering and tracking problems to project a complex posterior to an assumed simpler form (often the same form as the prior to maintain a closed-form).  In our case, we project the Beta mixture posterior to a single Beta in order to maintain a closed-form Beta approximation of the posterior update matching the form of the Beta prior in Eq~\eqref{prior}.

To apply ADF, we assume a Beta distribution with parameters $\alpha^{\prime}$ and $\beta^{\prime}$  for the posterior, and approximate the original mixture of Beta's with this distribution by equating their first moments (i.e., their means):
\begin{align}
    \frac{\alpha^{\prime}}{\alpha^{\prime}+\beta^{\prime}} & =\frac{w^{t}_{i}(1+\alpha)}{(1+\alpha+\beta)}+\frac{\alpha(1-w^{t}_{i})}{1+\alpha+\beta}\\
    & =\frac{\left[ {\alpha+w^{t}_{i}} \right]} {\left[ \alpha+w^{t}_{i} \right]
    +\left[ \beta+(1-w^{t}_{i}) \right] }.
\end{align}
Equating the numerators yields
\begin{equation}
    \alpha^{\prime} = \alpha+w^{t}_{i},
\end{equation}
and replacing this $\alpha^{\prime}$ in the equation of the denominators results in
\begin{equation}
        \beta^{\prime} = 1+\beta-w^{t}_{i}.
\end{equation}
Thus, the \textit{``mean matched''} posterior is $\text{Beta}(u_i;\alpha+w^{t}_{i}, 1+\beta-w^{t}_{i})$.

Matching two distributions by equating their first moments is a special case of a more general technique called \textit{``moment matching''}, which is widely used to approximate a complex probability distribution with a simpler one by equating their moments. In our work, we adopted a special case of this approach by matching the first moments, which we refer to as \textit{``mean matching''} of the distributions that we used for its simplicity and intuitive interpretation. However, this is only one of the possible solutions, and a complete moment matching derivation results in a slightly different solution. 

With this \textit{``mean matching''} derivation and current item $i$ posterior $Beta(u_i;\alpha,\beta)$ at step $t-1$, we can now perform an incremental posterior update after at step $t$ given the probability $w_i^t$ that the item description $x_i$ entails preference query $q_t$ yielding the closed-form Beta posterior $Beta(u_i;\alpha+w_i^t,1+\beta-w_i^t)$ as used in PEBOL-P.

\begin{table*}[h]
\centering
\caption{Examples of LLM Aspect-Based Query Generation from an Item Description}
\label{tab:llm_aspect_query}
\small 
\begin{tabular}{|p{1.5cm}|p{6.5cm}|p{2.5cm}|p{4cm}|} 
\hline
\textbf{Dataset} & \textbf{Item Description $d_i$} & \textbf{Generated Aspect $a^t$} & \textbf{Generated Query $q^t$} \\ \hline
\multirow{2}{*}{MovieLens} & Movie Title: Meet John Doe (1941) \newline Genres: Comedy, Drama \newline Tags: Christianity, Frank Capra, acting, anti-fascism, class issues, journalism, patriotic, pro american, thought provoking, AFI 100 (Cheers), BD-R, Barbara Stanwyck, Gary Cooper, baseball player, compare: This Is Our Land (2017), domain, funny, radio broadcast, reviewed in the NYer by Anthony Lanne (2018-04-30), suicide note & patriotism & Are you interested in movies with patriotic themes? \\ \cline{3-4} 
 & & classic & Do you enjoy classic movies? \\ \hline
\multirow{2}{*}{Recipe-MPR} & Spaghetti with mushrooms, onion, green pepper, chicken breasts, and alfredo sauce & alfredo sauce & Do you like alfredo sauce? \\ \cline{3-4} 
 & & chicken breast & Do you like chicken breasts? \\ \hline
\multirow{2}{*}{Yelp} & \begin{minipage}[t]{4.5cm}name: Le Pain Quotidien \\ categories: Restaurants, Bakeries, Breakfast \& Brunch, Coffee \& Tea, Food, Belgian, French\end{minipage} & bakery & Do you like bakeries? \\ \cline{3-4} 
 & & French pastries & Do you like French pastries? \\ \hline
\end{tabular}
\end{table*}


\begin{figure*} [ht]
    \centering
    \includegraphics[width=1\linewidth]{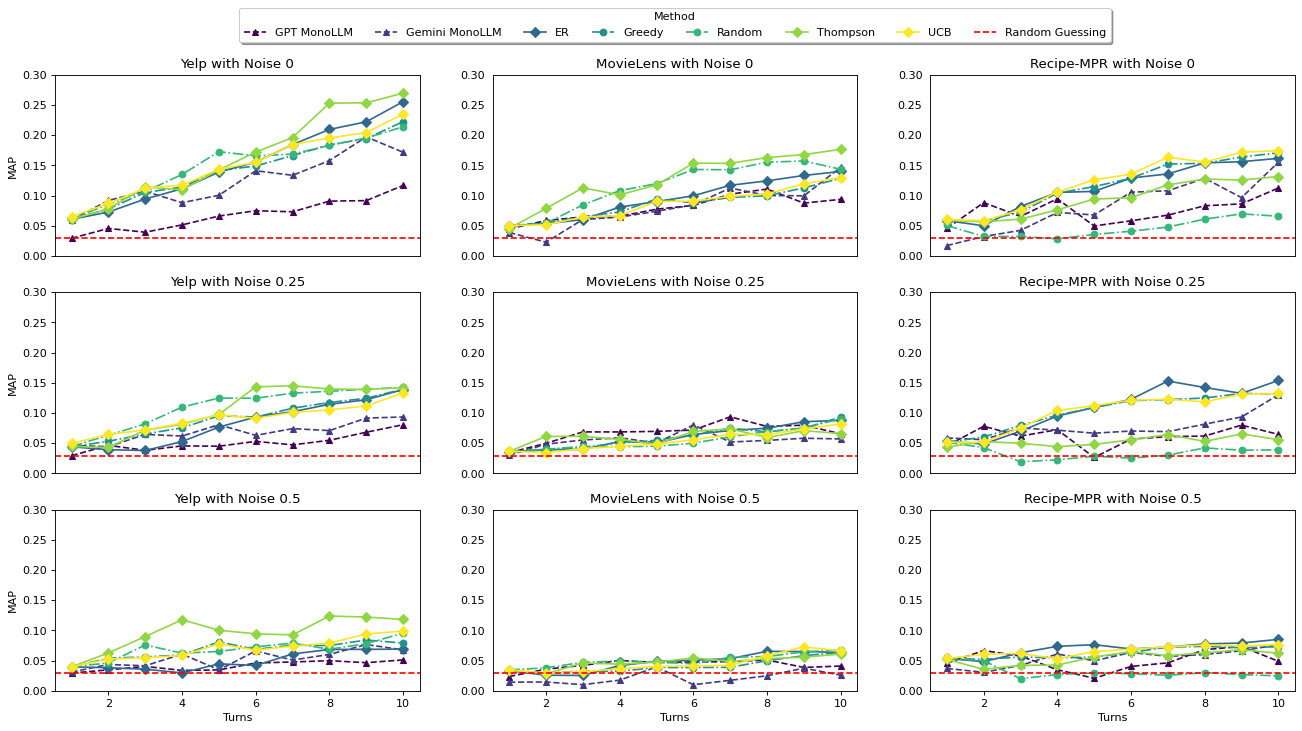}
    \caption{MAP@10 for all turns on all datasets and noise levels}
    \label{fig:mlni_full_noise_maps}
\end{figure*}

\begin{figure*}[ht] 
            \centering
            \includegraphics[width=1\linewidth]{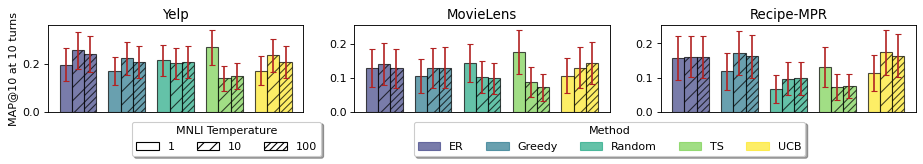}
            \caption{The effect of MNLI temperature on MAP@10 without noise -- error bars are 95\% C.I.s. Other results are reported using the temperatures that perform best across the most datasets for each policy.}
\label{fig:mnli_temp_comparison_noise0}
\end{figure*}

\begin{figure*} [ht]
    \centering
    \includegraphics[width=1\linewidth]{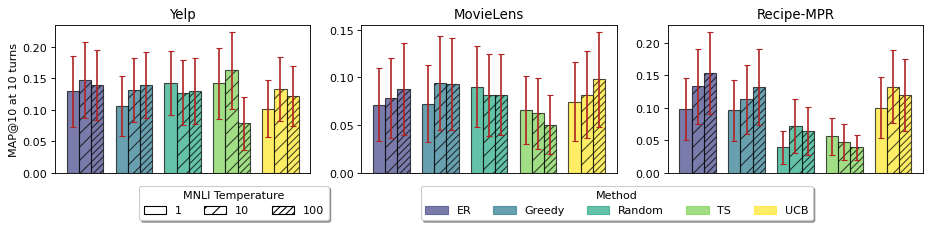}
    \caption{Effect of MNLI Temperature with noise 0.25}
    \label{fig:mlni_temp_noise0p25}
\end{figure*}

\begin{figure*}[ht]
    \centering
    \includegraphics[width=1\linewidth]{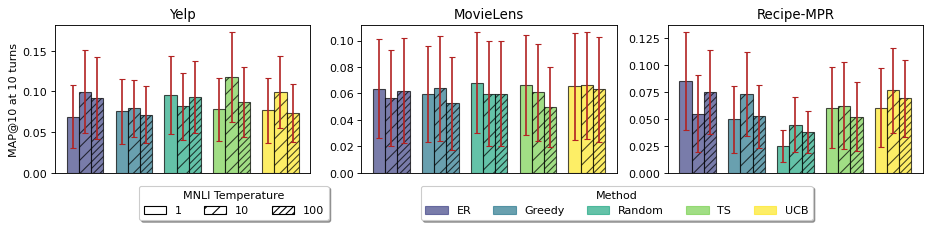}
    \caption{Effect of MNLI Temperature with noise 0.5}
    \label{fig:mnli_temp_noise0p5}
\end{figure*}

\begin{figure*}[t!]
        \subfloat[Monolithic LLM query generation]{%
            \includegraphics[width=.32\linewidth]{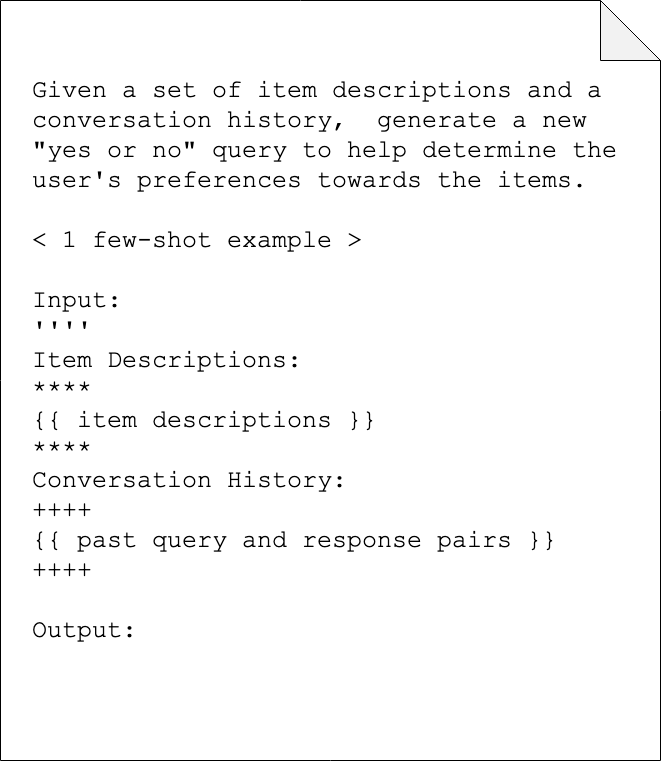}%
            \label{subfig:mono_query_gen_prompt}%
        }\hfill
        \subfloat[Monolithic LLM recommendation generation]{%
            \includegraphics[width=.32\linewidth]{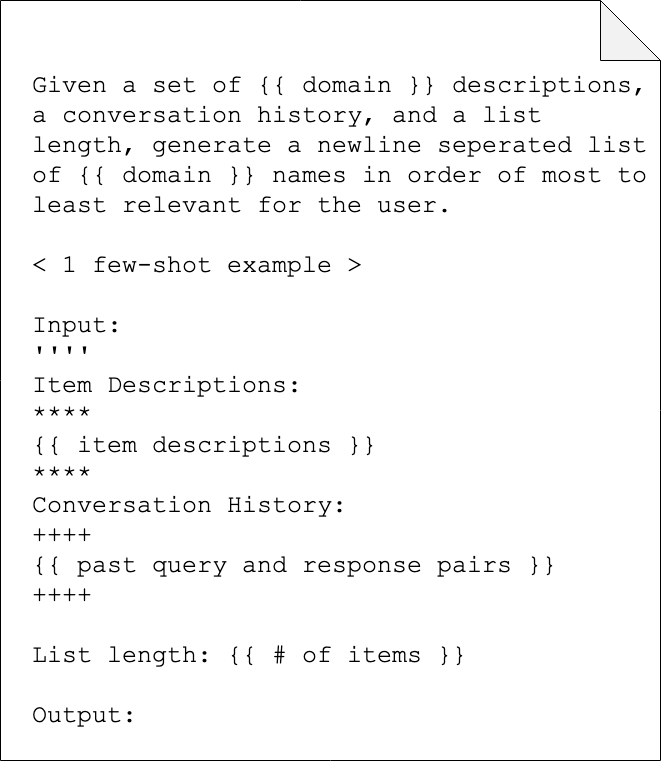}%
            \label{subfig:mono_recs_prompt}%
        }\hfill
        \subfloat[Aspect generation]{%
            \includegraphics[width=.32\linewidth]{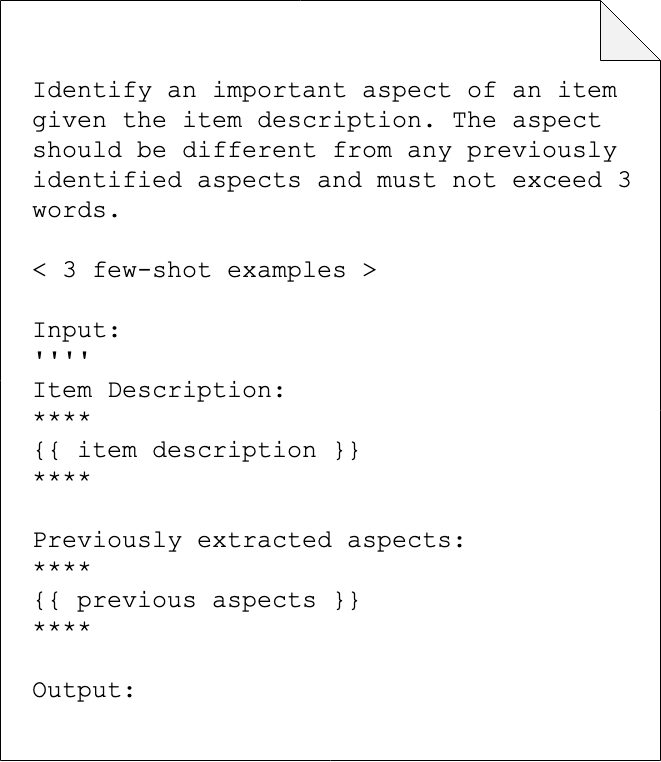}%
            \label{subfig:aspect_gen_prompt}%
        }\hfill
        \subfloat[Aspect-based query generation]{%
            \includegraphics[width=.32\linewidth]{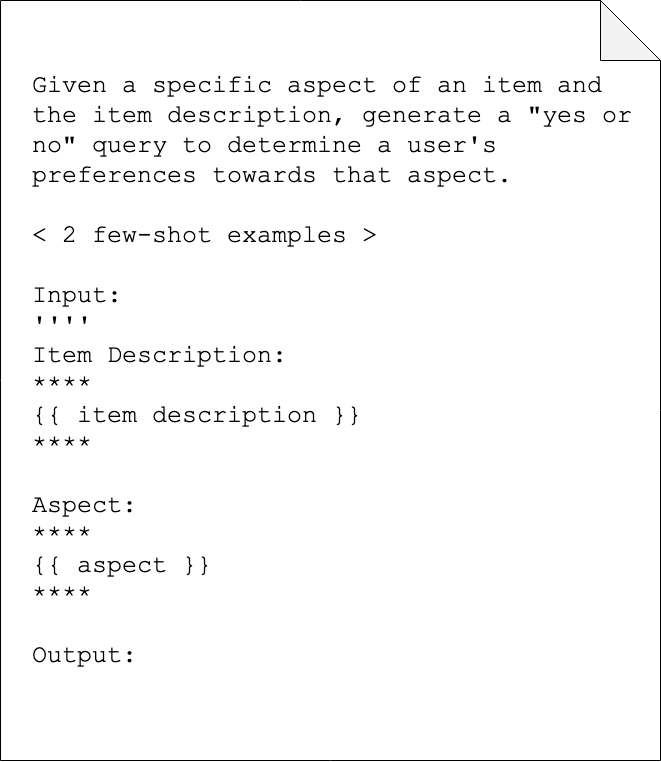}%
            \label{subfig:query_gen_prompt}%
        }
        \subfloat[User simulation]{%
            \includegraphics[width=.32\linewidth]{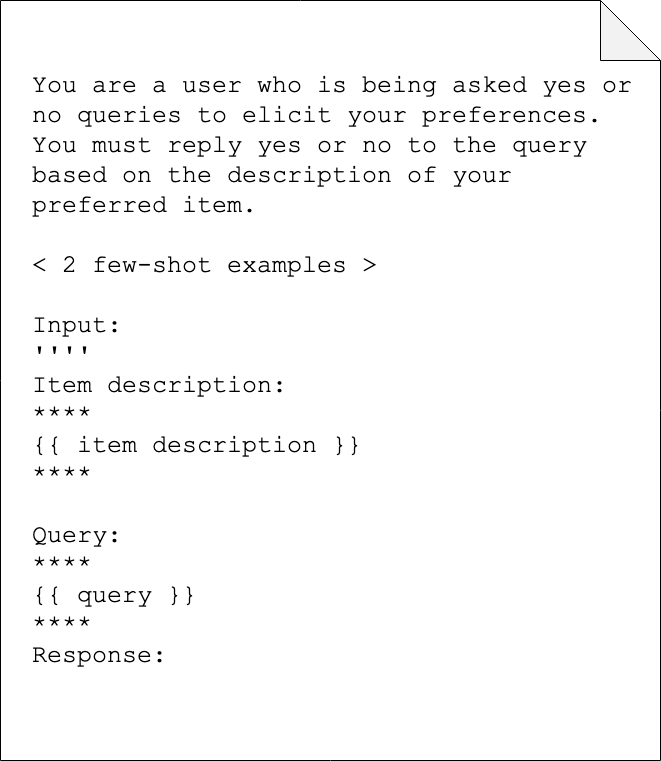}%
            \label{subfig:user_sim_prompt}%
        }\hfill
        \caption{LLM Prompting Templates}
        \label{fig:prompt_templates}
\end{figure*}


\begin{table*}[h]
\centering
\caption{MRR@10 Mean and 95\% C.I. for PEBOL-P vs MonoLLM on MovieLens without Response Noise}
\label{tab:}
\small 
\begin{tabular}{lrrrrrrrrrr}
\toprule
Turn & 1 & 2 & 3 & 4 & 5 & 6 & 7 & 8 & 9 & 10 \\
Method &  &  &  &  &  &  &  &  &  &  \\
\midrule
MonoLLM GPT Mean &       0.04 &       0.06 &       0.06 &       0.07 &       0.08 &       0.08 &       0.10 &       0.11 &       0.09 &       0.09 \\
MonoLLM GPT CI LB &       0.01 &       0.02 &       0.02 &       0.02 &       0.03 &       0.03 &       0.05 &       0.06 &       0.04 &       0.04 \\
MonoLLM GPT CI UB &       0.08 &       0.10 &       0.11 &       0.11 &       0.12 &       0.13 &       0.15 &       0.16 &       0.14 &       0.15 \\
MonoLLM Gemini Mean &       0.04 &       0.02 &       0.06 &       0.06 &       0.07 &       0.08 &       0.11 &       0.10 &       0.10 &       0.15 \\
MonoLLM Gemini CI LB &       0.01 &       0.01 &       0.02 &       0.03 &       0.04 &       0.05 &       0.07 &       0.05 &       0.06 &       0.09 \\
MonoLLM Gemini CI UB &       0.07 &       0.04 &       0.10 &       0.10 &       0.11 &       0.12 &       0.16 &       0.15 &       0.14 &       0.20 \\
\midrule
ER Mean &       \textbf{0.05} &       0.05 &       0.06 &       0.08 &       0.09 &       0.10 &       0.12 &       0.12 &       0.13 &       0.14 \\
ER CI LB &       0.01 &       0.02 &       0.03 &       0.04 &       0.05 &       0.05 &       0.06 &       0.07 &       0.07 &       0.08 \\
ER CI UB &       0.08 &       0.09 &       0.09 &       0.12 &       0.13 &       0.15 &       0.17 &       0.18 &       0.19 &       0.20 \\
Greedy Mean &       \textbf{0.05} &       0.05 &       0.07 &       0.07 &       0.09 &       0.09 &       0.10 &       0.10 &       0.11 &       0.13 \\
Greedy CI LB &       0.02 &       0.02 &       0.03 &       0.03 &       0.05 &       0.04 &       0.05 &       0.05 &       0.06 &       0.07 \\
Greedy CI UB &       0.09 &       0.09 &       0.10 &       0.11 &       0.14 &       0.14 &       0.15 &       0.15 &       0.17 &       0.19 \\
Random Mean &       \textbf{0.05} &       0.05 &       0.08 &       0.11 &       0.12 &       0.14 &       0.14 &       0.16 &       0.16 &       0.14 \\
Random CI LB &       0.01 &       0.02 &       0.04 &       0.05 &       0.06 &       0.08 &       0.08 &       0.09 &       0.09 &       0.09 \\
Random CI UB &       0.08 &       0.09 &       0.13 &       0.16 &       0.18 &       0.21 &       0.20 &       0.22 &       0.22 &       0.20 \\
TS Mean &       \textbf{0.05} &       \textbf{0.08} &       \textbf{0.11} &       \textbf{0.10} &       \textbf{0.12} &       \textbf{0.15} &       \textbf{0.15} &       \textbf{0.16} &       \textbf{0.17 }&       \textbf{0.18} \\
TS CI LB &       0.01 &       0.03 &       0.06 &       0.05 &       0.06 &       0.09 &       0.09 &       0.10 &       0.10 &       0.11 \\
TS CI UB &       0.08 &       0.13 &       0.17 &       0.15 &       0.17 &       0.22 &       0.21 &       0.23 &       0.23 &       0.24 \\
UCB Mean &       \textbf{0.05} &       0.05 &       0.06 &       0.07 &       0.09 &       0.09 &       0.10 &       0.10 &       0.12 &       0.13 \\
UCB CI LB &       0.02 &       0.02 &       0.03 &       0.03 &       0.05 &       0.04 &       0.05 &       0.05 &       0.06 &       0.07 \\
UCB CI UB &       0.09 &       0.09 &       0.10 &       0.10 &       0.13 &       0.14 &       0.15 &       0.15 &       0.17 &       0.19 \\
\bottomrule
\end{tabular}
\end{table*}

\begin{table*}[h]
\centering
\caption{MRR@10 Mean and 95\% C.I. for PEBOL-P vs MonoLLM on Recipe-MPR without Response Noise}
\label{tab:}
\small 
\begin{tabular}{lrrrrrrrrrr}
\toprule
Turn & 1 & 2 & 3 & 4 & 5 & 6 & 7 & 8 & 9 & 10 \\
Method &  &  &  &  &  &  &  &  &  &  \\
\midrule
MonoLLM GPT Mean &       0.05 &       \textbf{0.09} &       0.07 &       0.09 &       0.05 &       0.06 &       0.07 &       0.08 &       0.09 &       0.11 \\
MonoLLM GPT CI LB &       0.01 &       0.05 &       0.03 &       0.05 &       0.02 &       0.02 &       0.03 &       0.04 &       0.04 &       0.06 \\
MonoLLM GPT CI UB &       0.08 &       0.13 &       0.10 &       0.14 &       0.08 &       0.10 &       0.11 &       0.13 &       0.13 &       0.17 \\
MonoLLM Gemini Mean &       0.02 &       0.03 &       0.04 &       0.07 &       0.07 &       0.11 &       0.11 &       0.13 &       0.10 &       0.16 \\
MonoLLM Gemini CI LB &       0.01 &       0.02 &       0.02 &       0.04 &       0.03 &       0.06 &       0.07 &       0.08 &       0.06 &       0.10 \\
MonoLLM Gemini CI UB &       0.03 &       0.05 &       0.07 &       0.10 &       0.10 &       0.15 &       0.15 &       0.18 &       0.13 &       0.21 \\
\midrule
ER Mean &       \textbf{0.06} &       0.05 &       \textbf{0.08} &       \textbf{0.11} &       0.11 &       0.13 &       0.14 &       0.15 &       0.16 &       0.16 \\
ER CI LB &       0.02 &       0.02 &       0.04 &       0.06 &       0.06 &       0.08 &       0.08 &       0.09 &       0.09 &       0.10 \\
ER CI UB &       0.10 &       0.08 &       0.13 &       0.16 &       0.16 &       0.18 &       0.19 &       0.22 &       0.22 &       0.22 \\
Greedy Mean &       \textbf{0.06} &       0.06 &       0.07 &       0.10 &       0.11 &       0.13 &       0.15 &       0.15 &       0.16 &       0.17 \\
Greedy CI LB &       0.02 &       0.02 &       0.03 &       0.05 &       0.06 &       0.07 &       0.09 &       0.09 &       0.10 &       0.11 \\
Greedy CI UB &       0.10 &       0.09 &       0.11 &       0.15 &       0.17 &       0.18 &       0.21 &       0.21 &       0.23 &       0.24 \\
Random Mean &       0.05 &       0.03 &       0.03 &       0.03 &       0.04 &       0.04 &       0.05 &       0.06 &       0.07 &       0.07 \\
Random CI LB &       0.02 &       0.01 &       0.01 &       0.00 &       0.01 &       0.01 &       0.02 &       0.02 &       0.03 &       0.02 \\
Random CI UB &       0.08 &       0.06 &       0.06 &       0.05 &       0.06 &       0.07 &       0.08 &       0.10 &       0.11 &       0.11 \\
TS Mean &       \textbf{0.06} &       0.06 &       0.06 &       0.08 &       0.09 &       0.10 &       0.12 &       0.13 &       0.13 &       0.13 \\
TS CI LB &       0.02 &       0.02 &       0.02 &       0.03 &       0.05 &       0.05 &       0.06 &       0.07 &       0.07 &       0.07 \\
TS CI UB &       0.10 &       0.09 &       0.10 &       0.12 &       0.14 &       0.14 &       0.17 &       0.19 &       0.18 &       0.19 \\
UCB Mean &       \textbf{0.06} &       0.06 &       \textbf{0.08} &       \textbf{0.11} &       \textbf{0.13} &       \textbf{0.14} &       \textbf{0.16} &       \textbf{0.16} &       \textbf{0.17} &       \textbf{0.17} \\
UCB CI LB &       0.02 &       0.02 &       0.03 &       0.06 &       0.07 &       0.08 &       0.10 &       0.09 &       0.11 &       0.11 \\
UCB CI UB &       0.10 &       0.09 &       0.12 &       0.16 &       0.18 &       0.19 &       0.23 &       0.22 &       0.24 &       0.24 \\
\bottomrule
\end{tabular}
\end{table*}

\begin{table*}[h]
\centering
\caption{MRR@10 Mean and 95\% C.I. for PEBOL-P vs MonoLLM on Yelp without Response Noise}
\label{tab:}
\small 
\begin{tabular}{lrrrrrrrrrr}
\toprule
Turn &  1 & 2 & 3 & 4 & 5 & 6 & 7 & 8 & 9 & 10 \\
Method &  &  &  &  &  &  &  &  &  &  \\
\midrule
MonoLLM GPT Mean &       0.03 &       0.05 &       0.04 &       0.05 &       0.07 &       0.07 &       0.07 &       0.09 &       0.09 &       0.12 \\
MonoLLM GPT CI LB &       0.01 &       0.01 &       0.01 &       0.02 &       0.03 &       0.03 &       0.03 &       0.04 &       0.04 &       0.06 \\
MonoLLM GPT CI UB &       0.05 &       0.08 &       0.07 &       0.08 &       0.10 &       0.12 &       0.12 &       0.14 &       0.14 &       0.17 \\
MonoLLM Gemini Mean &       0.06 &       \textbf{0.09} &       \textbf{0.11} &       0.09 &       0.10 &       0.14 &       0.13 &       0.16 &       0.20 &       0.17 \\
MonoLLM Gemini CI LB &       0.02 &       0.05 &       0.06 &       0.05 &       0.06 &       0.08 &       0.09 &       0.10 &       0.14 &       0.12 \\
MonoLLM Gemini CI UB &       0.10 &       0.14 &       0.15 &       0.13 &       0.14 &       0.20 &       0.18 &       0.21 &       0.26 &       0.23 \\
\midrule
ER Mean &       0.06 &       0.07 &       0.09 &       0.11 &       0.14 &       0.16 &       0.18 &       0.21 &       0.22 &       0.26 \\
ER CI LB &       0.03 &       0.03 &       0.05 &       0.06 &       0.08 &       0.09 &       0.12 &       0.14 &       0.15 &       0.18 \\
ER CI UB &       0.10 &       0.11 &       0.14 &       0.16 &       0.20 &       0.22 &       0.25 &       0.28 &       0.29 &       0.33 \\
Greedy Mean &       0.06 &       0.08 &       \textbf{0.11} &       0.11 &       0.14 &       0.15 &       0.17 &       0.18 &       0.19 &       0.22 \\
Greedy CI LB &       0.03 &       0.04 &       0.06 &       0.07 &       0.08 &       0.09 &       0.10 &       0.12 &       0.13 &       0.16 \\
Greedy CI UB &       0.10 &       0.12 &       0.15 &       0.16 &       0.20 &       0.21 &       0.23 &       0.25 &       0.26 &       0.29 \\
Random Mean &       0.06 &       \textbf{0.09} &       \textbf{0.11} &       \textbf{0.13} &       \textbf{0.17} &       0.17 &       0.17 &       0.18 &       0.20 &       0.21 \\
Random CI LB &       0.03 &       0.04 &       0.06 &       0.08 &       0.11 &       0.11 &       0.11 &       0.12 &       0.13 &       0.15 \\
Random CI UB &       0.10 &       0.13 &       0.16 &       0.19 &       0.23 &       0.23 &       0.23 &       0.24 &       0.26 &       0.28 \\
TS Mean &       0.06 &       0.08 &       \textbf{0.11} &       0.11 &       0.14 &       \textbf{0.17} &       \textbf{0.20} &       \textbf{0.25} &       \textbf{0.25} &       \textbf{0.27} \\
TS CI LB &       0.03 &       0.04 &       0.06 &       0.07 &       0.09 &       0.11 &       0.13 &       0.18 &       0.18 &       0.20 \\
TS CI UB &       0.10 &       0.12 &       0.17 &       0.15 &       0.20 &       0.23 &       0.26 &       0.33 &       0.32 &       0.34 \\
UCB Mean &       \textbf{0.07} &       \textbf{0.09} &       \textbf{0.11} &       0.12 &       0.14 &       0.16 &       0.18 &       0.20 &       0.20 &       0.23 \\
UCB CI LB &       0.03 &       0.05 &       0.06 &       0.07 &       0.08 &       0.09 &       0.12 &       0.13 &       0.14 &       0.16 \\
UCB CI UB &       0.10 &       0.13 &       0.16 &       0.17 &       0.20 &       0.22 &       0.25 &       0.26 &       0.27 &       0.30 \\
\bottomrule
\end{tabular}
\end{table*}

\begin{table*}[h]
\centering
\caption{MRR@10 Mean and 95\% C.I. for PEBOL-P without Aspect History on MovieLens without Response Noise}
\label{tab:}
\small 
\begin{tabular}{lrrrrrrrrrr}
\toprule
Turn & 1 & 2 & 3 & 4 & 5 & 6 & 7 & 8 & 9 & 10 \\
Method &  &  &  &  &  &  &  &  &  &  \\
\midrule
TS Mean &       \textbf{0.03} &       \textbf{0.05} &       \textbf{0.06} &       \textbf{0.06} &       \textbf{0.08} &       \textbf{0.11} &       \textbf{0.13} &       \textbf{0.12} &       \textbf{0.13} &       \textbf{0.14} \\
TS CI LB &       0.01 &       0.02 &       0.02 &       0.02 &       0.04 &       0.06 &       0.08 &       0.06 &       0.07 &       0.08 \\
TS CI UB &       0.06 &       0.09 &       0.09 &       0.09 &       0.13 &       0.16 &       0.19 &       0.17 &       0.18 &       0.19 \\
UCB Mean &       \textbf{0.03} &       0.03 &       0.04 &       0.05 &       0.05 &       0.05 &       0.06 &       0.06 &       0.06 &       0.06 \\
UCB CI LB &       0.01 &       0.01 &       0.02 &       0.02 &       0.02 &       0.02 &       0.02 &       0.02 &       0.02 &       0.02 \\
UCB CI UB &       0.05 &       0.06 &       0.07 &       0.09 &       0.08 &       0.08 &       0.09 &       0.10 &       0.09 &       0.09 \\
\bottomrule
\end{tabular}
\end{table*}

\begin{table*}[h]
\centering
\caption{MRR@10 Mean and 95\% C.I. for PEBOL-P without Aspect History on Recipe-MPR without Response Noise}
\label{tab:}
\small 
\begin{tabular}{lrrrrrrrrrr}
\toprule
Turn & 1 & 2 & 3 & 4 & 5 & 6 & 7 & 8 & 9 & 10 \\
Method &  &  &  &  &  &  &  &  &  &  \\
\midrule
TS Mean &       \textbf{0.04} &       \textbf{0.05} &       0.03 &       0.04 &       0.06 &       0.07 &       0.06 &       \textbf{0.09} &       \textbf{0.10} &       0.09 \\
TS CI LB &       0.01 &       0.02 &       0.01 &       0.01 &       0.03 &       0.03 &       0.03 &       0.04 &       0.05 &       0.05 \\
TS CI UB &       0.07 &       0.09 &       0.05 &       0.07 &       0.10 &       0.12 &       0.10 &       0.13 &       0.14 &       0.14 \\
UCB Mean &       \textbf{0.04} &       \textbf{0.05} &       \textbf{0.06} &       \textbf{0.08} &       \textbf{0.08} &       \textbf{0.09} &      \textbf{ 0.08} &       \textbf{0.09} &      \textbf{ 0.10} &       \textbf{0.10} \\
UCB CI LB &       0.01 &       0.01 &       0.02 &       0.04 &       0.04 &       0.04 &       0.04 &       0.04 &       0.05 &       0.05 \\
UCB CI UB &       0.07 &       0.08 &       0.09 &       0.12 &       0.12 &       0.13 &       0.12 &       0.13 &       0.14 &       0.14 \\
\bottomrule
\end{tabular}
\end{table*}

\begin{table*}[h]
\centering
\caption{MRR@10 Mean and 95\% C.I. for PEBOL-P without Aspect History on Yelp without Response Noise}
\label{tab:}
\small 
\begin{tabular}{lrrrrrrrrrr}
\toprule
Turn & 1 & 2 & 3 & 4 & 5 & 6 & 7 & 8 & 9 & 10 \\
Method &  &  &  &  &  &  &  &  &  &  \\
\midrule
TS Mean &       \textbf{0.05} &       0.06 &       0.07 &       0.08 &       0.09 &       0.11 &       0.14 &       0.14 &       0.15 &       0.16 \\
TS CI LB &       0.02 &       0.02 &       0.03 &       0.04 &       0.05 &       0.06 &       0.09 &       0.08 &       0.09 &       0.10 \\
TS CI UB &       0.09 &       0.09 &       0.10 &       0.11 &       0.13 &       0.16 &       0.20 &       0.19 &       0.21 &       0.22 \\
UCB Mean &       \textbf{0.05} &       \textbf{0.08} &       \textbf{0.09} &       \textbf{0.11} &       \textbf{0.12} &       \textbf{0.15} &       \textbf{0.15} &       \textbf{0.17} &       \textbf{0.17} &       \textbf{0.19} \\
UCB CI LB &       0.02 &       0.04 &       0.05 &       0.06 &       0.07 &       0.09 &       0.10 &       0.11 &       0.11 &       0.13 \\
UCB CI UB &       0.09 &       0.12 &       0.13 &       0.16 &       0.17 &       0.20 &       0.21 &       0.22 &       0.22 &       0.25 \\
\bottomrule
\end{tabular}
\end{table*}

\begin{table*}[h]
\centering
\caption{MRR@10 Mean and 95\% C.I. for PEBOL-B on MovieLens without Response Noise}
\label{tab:}
\small 
\begin{tabular}{lrrrrrrrrrr}
\toprule
Turn & 1 & 2 & 3 & 4 & 5 & 6 & 7 & 8 & 9 & 10 \\
Method &  &  &  &  &  &  &  &  &  &  \\
\midrule
ER Mean &       \textbf{0.03} &       0.04 &       0.04 &       0.05 &       0.06 &       0.08 &       0.09 &       0.11 &       0.11 &       0.12 \\
ER CI LB &       0.01 &       0.01 &       0.01 &       0.02 &       0.03 &       0.04 &       0.04 &       0.06 &       0.07 &       0.07 \\
ER CI UB &       0.06 &       0.07 &       0.08 &       0.08 &       0.10 &       0.12 &       0.14 &       0.16 &       0.16 &       0.16 \\
Greedy Mean &       \textbf{0.03} &       0.04 &       0.05 &       0.05 &       0.07 &       \textbf{0.09} &       \textbf{0.10} &       0.11 &       0.11 &       0.11 \\
Greedy CI LB &       0.01 &       0.01 &       0.02 &       0.02 &       0.03 &       0.05 &       0.05 &       0.06 &       0.07 &       0.07 \\
Greedy CI UB &       0.06 &       0.08 &       0.08 &       0.09 &       0.11 &       0.13 &       0.14 &       0.16 &       0.16 &       0.16 \\
Random Mean &       \textbf{0.03} &       0.04 &       0.04 &       0.06 &       0.06 &       0.08 &       0.09 &       0.10 &       0.10 &       0.10 \\
Random CI LB &       0.01 &       0.01 &       0.01 &       0.03 &       0.03 &       0.04 &       0.05 &       0.06 &       0.06 &       0.06 \\
Random CI UB &       0.06 &       0.06 &       0.07 &       0.09 &       0.10 &       0.12 &       0.13 &       0.14 &       0.14 &       0.14 \\
TS Mean &       \textbf{0.03} &       \textbf{0.05} &       \textbf{0.06} &       \textbf{0.07} &       \textbf{0.08} &       0.07 &       0.09 &       \textbf{0.12} &       \textbf{0.12} &       \textbf{0.13} \\
TS CI LB &       0.01 &       0.02 &       0.02 &       0.03 &       0.03 &       0.03 &       0.04 &       0.07 &       0.07 &       0.07 \\
TS CI UB &       0.06 &       0.08 &       0.09 &       0.11 &       0.12 &       0.11 &       0.13 &       0.17 &       0.18 &       0.18 \\
UCB Mean &       \textbf{0.03} &       0.04 &       0.05 &       0.05 &       0.06 &       0.08 &       0.09 &       0.10 &       0.11 &       0.11 \\
UCB CI LB &       0.01 &       0.01 &       0.02 &       0.02 &       0.03 &       0.04 &       0.05 &       0.06 &       0.06 &       0.06 \\
UCB CI UB &       0.06 &       0.07 &       0.08 &       0.08 &       0.09 &       0.13 &       0.14 &       0.15 &       0.15 &       0.15 \\
\bottomrule
\end{tabular}
\end{table*}

\begin{table*}[h]
\centering
\caption{MRR@10 Mean and 95\% C.I. for PEBOL-B on Recipe-MPR without Response Noise}
\label{tab:}
\small 
\begin{tabular}{lrrrrrrrrrr}
\toprule
Turn & 1 & 2 & 3 & 4 & 5 & 6 & 7 & 8 & 9 & 10 \\
Method &  &  &  &  &  &  &  &  &  &  \\
\midrule
ER Mean &       \textbf{0.04} &       \textbf{0.06} &       \textbf{0.09} &       \textbf{0.07} &       0.09 &       0.09 &       0.10 &       \textbf{0.13} &       0.10 &       0.10 \\
ER CI LB &       0.00 &       0.02 &       0.05 &       0.03 &       0.04 &       0.04 &       0.05 &       0.07 &       0.05 &       0.05 \\
ER CI UB &       0.07 &       0.10 &       0.14 &       0.10 &       0.14 &       0.14 &       0.16 &       0.19 &       0.15 &       0.15 \\
Greedy Mean &       \textbf{0.04} &       0.04 &       0.06 &       \textbf{0.07} &       \textbf{0.10} &       \textbf{0.11} &       \textbf{0.12} &       0.12 &       0.12 &       \textbf{0.13} \\
Greedy CI LB &       0.00 &       0.01 &       0.02 &       0.03 &       0.05 &       0.06 &       0.06 &       0.06 &       0.06 &       0.07 \\
Greedy CI UB &       0.07 &       0.08 &       0.10 &       0.10 &       0.15 &       0.17 &       0.18 &       0.18 &       0.18 &       0.19 \\
Random Mean &       \textbf{0.04} &       0.03 &       0.03 &       0.05 &       0.04 &       0.04 &       0.05 &       0.04 &       0.05 &       0.06 \\
Random CI LB &       0.00 &       0.00 &       0.01 &       0.01 &       0.01 &       0.01 &       0.01 &       0.01 &       0.02 &       0.02 \\
Random CI UB &       0.07 &       0.06 &       0.05 &       0.09 &       0.07 &       0.08 &       0.08 &       0.07 &       0.09 &       0.11 \\
TS Mean &       \textbf{0.04} &       \textbf{0.06} &       0.06 &       \textbf{0.07} &       0.09 &       0.09 &       0.10 &       0.10 &       0.11 &       0.12 \\
TS CI LB &       0.00 &       0.02 &       0.02 &       0.03 &       0.05 &       0.04 &       0.05 &       0.05 &       0.06 &       0.06 \\
TS CI UB &       0.07 &       0.09 &       0.10 &       0.12 &       0.14 &       0.14 &       0.16 &       0.15 &       0.17 &       0.17 \\
UCB Mean &       \textbf{0.04} &       0.05 &       0.07 &       \textbf{0.07} &       0.09 &       0.11 &       \textbf{0.12} &       0.12 &       \textbf{0.13} &       \textbf{0.13} \\
UCB CI LB &       0.00 &       0.02 &       0.03 &       0.03 &       0.04 &       0.05 &       0.06 &       0.07 &       0.07 &       0.08 \\
UCB CI UB &       0.07 &       0.08 &       0.11 &       0.11 &       0.14 &       0.16 &       0.17 &       0.18 &       0.18 &       0.19 \\
\bottomrule
\end{tabular}
\end{table*}

\begin{table*}[h]
\centering
\caption{MRR@10 Mean and 95\% C.I. for PEBOL-B on Yelp without Response Noise}
\label{tab:}
\small 
\begin{tabular}{lrrrrrrrrrr}
\toprule
Turn & 1 & 2 & 3 & 4 & 5 & 6 & 7 & 8 & 9 & 10 \\
Method &  &  &  &  &  &  &  &  &  &  \\
\midrule
ER Mean &       \textbf{0.05} &       0.05 &       0.07 &       0.09 &       0.11 &       0.11 &       0.11 &       0.11 &       0.11 &       0.11 \\
ER CI LB &       0.02 &       0.02 &       0.04 &       0.05 &       0.07 &       0.07 &       0.07 &       0.06 &       0.06 &       0.06 \\
ER CI UB &       0.08 &       0.09 &       0.11 &       0.12 &       0.15 &       0.15 &       0.15 &       0.15 &       0.15 &       0.15 \\
Greedy Mean &       \textbf{0.05} &       \textbf{0.06} &       0.08 &       0.09 &       0.10 &       0.10 &       0.10 &       0.10 &       0.10 &       0.10 \\
Greedy CI LB &       0.02 &       0.02 &       0.04 &       0.05 &       0.06 &       0.06 &       0.06 &       0.06 &       0.06 &       0.06 \\
Greedy CI UB &       0.08 &       0.09 &       0.11 &       0.13 &       0.14 &       0.13 &       0.13 &       0.14 &       0.14 &       0.13 \\
Random Mean &       \textbf{0.05} &       \textbf{0.06} &       0.07 &       0.09 &       0.11 &       0.10 &       0.10 &       0.11 &       0.12 &       0.14 \\
Random CI LB &       0.02 &       0.03 &       0.03 &       0.05 &       0.06 &       0.06 &       0.06 &       0.07 &       0.07 &       0.09 \\
Random CI UB &       0.08 &       0.09 &       0.11 &       0.14 &       0.15 &       0.14 &       0.14 &       0.15 &       0.16 &       0.19 \\
TS Mean &       \textbf{0.05} &       \textbf{0.06} &       \textbf{0.12} &       \textbf{0.12} &       \textbf{0.16} &       \textbf{0.16} &       \textbf{0.18} &       \textbf{0.18} &       \textbf{0.22} &       \textbf{0.24} \\
TS CI LB &       0.02 &       0.02 &       0.07 &       0.07 &       0.10 &       0.10 &       0.12 &       0.12 &       0.15 &       0.17 \\
TS CI UB &       0.08 &       0.10 &       0.17 &       0.17 &       0.22 &       0.22 &       0.24 &       0.24 &       0.28 &       0.31 \\
UCB Mean &       \textbf{0.05} &       \textbf{0.06} &       0.08 &       0.10 &       0.10 &       0.10 &       0.10 &       0.10 &       0.10 &       0.10 \\
UCB CI LB &       0.02 &       0.02 &       0.04 &       0.06 &       0.06 &       0.06 &       0.06 &       0.06 &       0.06 &       0.06 \\
UCB CI UB &       0.08 &       0.09 &       0.12 &       0.13 &       0.14 &       0.14 &       0.14 &       0.14 &       0.14 &       0.13 \\
\bottomrule
\end{tabular}
\end{table*}

\begin{table*}[h]
\centering
\caption{MRR@10 Mean and 95\% C.I. at Turn 10 for PEBOL-P and MonoLLM baselines on MovieLens with Different Levels of Response Noise}
\label{tab:}
\small 
\begin{tabular}{lrrr}
\toprule
Noise Level & 0 & 0.25 & 0.5 \\
Method &  &  &  \\
\midrule
MonoLLM GPT Mean &       0.09 &       0.07 &       0.04 \\
MonoLLM GPT CI LB &       0.04 &       0.03 &       0.01 \\
MonoLLM GPT CI UB &       0.15 &       0.10 &       0.07 \\
MonoLLM Gemini Mean &       0.15 &       0.06 &       0.03 \\
MonoLLM Gemini CI LB &       0.09 &       0.02 &       0.00 \\
MonoLLM Gemini CI UB &       0.20 &       0.09 &       0.05 \\
\midrule
ER Mean &       0.14 &       \textbf{0.09} &       0.06 \\
ER CI LB &       0.08 &       0.04 &       0.03 \\
ER CI UB &       0.20 &       0.14 &       0.10 \\
Greedy Mean &       0.13 &       \textbf{0.09} &       0.06 \\
Greedy CI LB &       0.07 &       0.04 &       0.02 \\
Greedy CI UB &       0.19 &       0.14 &       0.10 \\
Random Mean &       0.14 &       \textbf{0.09} &       \textbf{0.07} \\
Random CI LB &       0.09 &       0.05 &       0.03 \\
Random CI UB &       0.20 &       0.13 &       0.11 \\
TS Mean &       \textbf{0.18} &       0.07 &       0.06 \\
TS CI LB &       0.11 &       0.03 &       0.02 \\
TS CI UB &       0.24 &       0.10 &       0.10 \\
UCB Mean &       0.13 &       0.08 &       \textbf{0.07} \\
UCB CI LB &       0.07 &       0.04 &       0.03 \\
UCB CI UB &       0.19 &       0.13 &       0.11 \\
\bottomrule
\end{tabular}
\end{table*}

\begin{table*}[h]
\centering
\caption{MRR@10 Mean and 95\% C.I. at Turn 10 for PEBOL-P and MonoLLM baselines on Recipe-MPR with Different Levels of Response Noise}
\label{tab:}
\small 
\begin{tabular}{lrrr}
\toprule
Noise Level & 0 & 0.25 & 0.5 \\
Method &  &  &  \\
\midrule
MonoLLM GPT Mean &       0.11 &       0.06 &       0.05 \\
MonoLLM GPT CI LB &       0.06 &       0.02 &       0.01 \\
MonoLLM GPT CI UB &       0.17 &       0.11 &       0.08 \\
MonoLLM Gemini Mean &       0.16 &       0.13 &       0.08 \\
MonoLLM Gemini CI LB &       0.10 &       0.07 &       0.03 \\
MonoLLM Gemini CI UB &       0.21 &       0.19 &       0.12 \\
\midrule
ER Mean &       0.16 &       \textbf{0.15} &       \textbf{0.09} \\
ER CI LB &       0.10 &       0.09 &       0.04 \\
ER CI UB &       0.22 &       0.22 &       0.13 \\
Greedy Mean &       \textbf{0.17} &       0.13 &       0.07 \\
Greedy CI LB &       0.11 &       0.07 &       0.03 \\
Greedy CI UB &       0.24 &       0.19 &       0.11 \\
Random Mean &       0.07 &       0.04 &       0.02 \\
Random CI LB &       0.02 &       0.01 &       0.01 \\
Random CI UB &       0.11 &       0.06 &       0.04 \\
TS Mean &       0.13 &       0.06 &       0.06 \\
TS CI LB &       0.07 &       0.03 &       0.02 \\
TS CI UB &       0.19 &       0.08 &       0.10 \\
UCB Mean &       \textbf{0.17} &       0.13 &       0.08 \\
UCB CI LB &       0.11 &       0.08 &       0.04 \\
UCB CI UB &       0.24 &       0.19 &       0.12 \\
\bottomrule
\end{tabular}
\end{table*}

\begin{table*}[h]
\centering
\caption{MRR@10 Mean and 95\% C.I. at Turn 10 for PEBOL-P and MonoLLM baselines on Yelp with Different Levels of Response Noise}
\label{tab:}
\small 
\begin{tabular}{lrrr}
\toprule
Noise Level & 0 & 0.25 & 0.5 \\
Method &  &  &  \\
\midrule
MonoLLM GPT Mean &       0.12 &       0.08 &       0.05 \\
MonoLLM GPT CI LB &       0.06 &       0.03 &       0.01 \\
MonoLLM GPT CI UB &       0.17 &       0.13 &       0.09 \\
MonoLLM Gemini Mean &       0.17 &       0.09 &       0.07 \\
MonoLLM Gemini CI LB &       0.12 &       0.05 &       0.02 \\
MonoLLM Gemini CI UB &       0.23 &       0.14 &       0.11 \\
\midrule
ER Mean &       0.26 &       \textbf{0.14} &       0.07 \\
ER CI LB &       0.18 &       0.08 &       0.03 \\
ER CI UB &       0.33 &       0.19 &       0.11 \\
Greedy Mean &       0.22 &       \textbf{0.14} &       0.08 \\
Greedy CI LB &       0.16 &       0.09 &       0.04 \\
Greedy CI UB &       0.29 &       0.19 &       0.11 \\
Random Mean &       0.21 &       \textbf{0.14} &       0.10 \\
Random CI LB &       0.15 &       0.09 &       0.05 \\
Random CI UB &       0.28 &       0.19 &       0.14 \\
TS Mean &       \textbf{0.27} &       \textbf{0.14} &       \textbf{0.12} \\
TS CI LB &       0.20 &       0.08 &       0.06 \\
TS CI UB &       0.34 &       0.20 &       0.17 \\
UCB Mean &       0.23 &       0.13 &       0.10 \\
UCB CI LB &       0.16 &       0.08 &       0.05 \\
UCB CI UB &       0.30 &       0.18 &       0.14 \\
\bottomrule
\end{tabular}
\end{table*}

\end{document}